\documentclass{ieeeaccess}
\usepackage{cite}
\usepackage{amsmath,amssymb,amsfonts}
\usepackage{algorithmic}
\usepackage{graphicx}
\usepackage{subcaption}
\usepackage{booktabs}
\usepackage{textcomp}
\usepackage{multirow}

\usepackage{soul}
\usepackage{pifont}
\newcommand{\cmark}{\ding{51}}%
\newcommand{\xmark}{\ding{55}}
\def\BibTeX{{\rm B\kern-.05em{\sc i\kern-.025em b}\kern-.08em
    T\kern-.1667em\lower.7ex\hbox{E}\kern-.125emX}}
\begin{document}
\history{Date of publication xxxx 00, 0000, date of current version xxxx 00, 0000.}
\doi{10.1109/ACCESS.2024.3356913}

\title{CasTGAN: Cascaded Generative Adversarial Network for Realistic Tabular Data Synthesis}
\author{
\uppercase{Abdallah Alshantti}\authorrefmark{1,2},
\uppercase{Damiano Varagnolo}\authorrefmark{1},
\uppercase{Adil Rasheed}\authorrefmark{1},
\uppercase{Aria Rahmati}\authorrefmark{3},
and \uppercase{Frank Westad}\authorrefmark{1}
}
\address[1]{Department of Engineering Cybernetics, Norwegian University of Science and Technology, 7034 Trondheim, Norway}
\address[2]{DNB ASA, 0131 Oslo, Norway}
\address[3]{Current affiliation: Sopra Steria, 0185 Oslo, Norway. Contribution made while author was employed at DNB ASA.}
\tfootnote{This work was supported by DNB ASA through the funding of this research project.}

\markboth
{A. Alshantti \headeretal: CasTGAN: Cascaded Generative Adversarial Network for Realistic Tabular Data Synthesis}
{A. Alshantti \headeretal: CasTGAN: Cascaded Generative Adversarial Network for Realistic Tabular Data Synthesis}

\corresp{Corresponding author: Abdallah Alshantti (e-mail: abdallah.a.s.alshantti@ntnu.no).}

\begin{abstract}
Generative adversarial networks (GANs) have drawn considerable attention in recent years for their proven capability in generating synthetic data which can be utilised for multiple purposes. While GANs have demonstrated tremendous successes in producing synthetic data samples that replicate the dynamics of the original datasets, the validity of the synthetic data and the underlying privacy concerns represent major challenges which are not sufficiently addressed. In this work, we design a cascaded tabular GAN framework (CasTGAN) for generating realistic tabular data with a specific focus on the validity of the output. In this context, validity refers to the the dependency between features that can be found in the real data, but is typically misrepresented by traditional generative models. Our key idea entails that employing a cascaded architecture in which a dedicated generator samples each feature, the synthetic output becomes more representative of the real data. Our experimental results demonstrate that our model is capable of generating synthetic tabular data that can be used for fitting machine learning models, as CasTGAN's classification performance only falls under the real training data's PR-AUC score by 4.88\% on average for classification datasets, and exhibits an average reduction of the real training data's $R^2$ score by 0.139 for regression datasets. In addition, our model captures well the constraints and the correlations between the features of the real data, especially the high dimensional datasets. Assessing the generation of invalid records demonstrates that CasTGAN reduces the number of invalid data observations by up to 622\% in comparison to the second best performing baseline tabular GAN model. Furthermore, we evaluate the risk of white-box privacy attacks on our model and subsequently show that applying some perturbations to the auxiliary learners in CasTGAN increases the overall robustness of our model against targeted attacks.
\end{abstract}

\begin{keywords}
Generative adversarial networks, output validity, privacy attacks, tabular data.
\end{keywords}

\titlepgskip=-21pt

\maketitle

\section{Introduction}
\label{sc:intro}
Facilitating information and knowledge sharing within and between organisations is increasingly sought after for attaining growth and development. From a healthcare and medical standpoint, information exchange subsequently contributes to better understanding of diseases and risk factors, more intuitive prognosis by practitioners and effective treatment planning based on previously obtained knowledge \cite{walonoski2018synthea}. In the financial sector, sharing information between stakeholders leads to improved prediction of corporate bankruptcy and quicker identification of suspicious transaction behaviour that can be potentially linked to organised financial crime \cite{pavlidis2020financial}. For both fields, sharing the data which contains sensitive patient and client information is subject to the European Union's General Data Protection Regulation (GDPR) \cite{regulation2016regulation} to maintain the confidentiality and privacy of such information. Therefore, institutions are continuously seeking new data anonymisation and synthetic data generation techniques for exchanging domain knowledge without exposing sensitive information.

Ever since their development, generative adversarial networks (GANs) \cite{goodfellow2014generative} have been increasingly studied for their ability to approximate and model complex data distributions. Despite the early GAN applications being densely focused on the computer vision domain and image generation, GANs are becoming recently researched in other fields such as natural language processing \cite{young2018recent} and time-series anomaly detection \cite{li2019mad}. In addition, more properties of GANs have emerged such as conditionally generating samples based on a specific target class \cite{mirza2014conditional} and generation in conjunction with variational auto-encoders \cite{makhzani2015adversarial}. 

In contrast, GANs have been significantly less explored for tabular data generation. A tabular dataset typically comprises a mixture of continuous variables and categorical features \cite{esmaeilpour2022bi}. Tabular data is common in the medical and the financial domains where fields such age, gender, profession, and income can be commonly found in databases containing numerous records. As opposed to purely numerical data, representing datasets with categorical variables can be particularly difficult in presence of highly-dimensional and strongly correlated features. Furthermore, quantifying the validity of a synthetic tabular dataset can be practically impossible without closely inspecting every generated data sample and deciding whether to accept or reject each examined data record. Correspondingly, invalidity refers to the semantically incorrect representation of the features, where the interdependencies between some data features are not correctly modelled. This constitutes a challenge when synthetic data is harnessed for knowledge exchange, as semantically incorrect data can lead to misinterpretation and flawed understanding of the data, hence disparaging the effect of facilitating knowledge sharing. Notwithstanding, there currently exists no straightforward and unified criteria for evaluating the validity of the output generated by tabular GANs \cite{assefa2020generating}.

To rectify the previously outlined limitations, we introduce CasTGAN, which is a generative network framework characterised by multiple generators connected sequentially; each of which is designed to generate a single feature. Meanwhile, a single discriminator validates the output of all the generators while being trained on the output of the final generator in the cascade. In addition, each generator is chained to a corresponding auxiliary learner in order to obtain more insightful losses specific to the individually generated features. This is motivated by the fact that it has been shown that adding more auxiliary classifiers can enhance the quality of the synthetic output images \cite{odena2017conditional}. Therefore, we posit that CasTGAN aims to capture the highly correlated and hierarchical relationship between features, such that the synthetic output produced by our model closely resembles the real data while minimising the inconsistencies in the generated data. This is particularly important for applications where data is widely shared between professionals, and the slightest irregularities in the data can lead to undesired outcomes.

We can thereby summarise our contributions in this work as: 
\begin{itemize}
    \item \emph{Generative architecture}: A cascaded based generative framework for producing realistic tabular output which greatly emulates the original data, while significantly reducing the number of invalid synthetic samples.
    \item \emph{Synthetic data evaluation}: A new metric for quantifying the \emph{realistic-ness} of the synthetic data when lacking the domain knowledge for the provided data, and extensively evaluate our framework and existing works. 
    \item \emph{Privacy assessment}: We launch white-box privacy attacks on our model and analyse how the privacy guarantees and quality of the output are impacted when perturbing the input data during the model training.  
\end{itemize}

The remainder of this paper is structured as follows. In Section \ref{sc:background}, we present an overview of GANs and the types of GAN privacy attacks, while we further examine the relevant studies in Section \ref{sc:related_works}. Section \ref{sc:methodology} presents a discussion of CasTGAN and a detailed description of our model's structure. In Section \ref{sc:exp_setup}, we outline the experimental setup used in this work and the evaluation criteria. We demonstrate our results in Section \ref{sc:results} and discuss our findings in Section \ref{sc:discussion}. The paper is concluded in Section \ref{sc:conclusion}.


\section{Background}
\label{sc:background}

\subsection{Generative Adversarial Networks}

A GAN is characterised by a generator $G$ and a discriminator $D$ playing an adversarial game, where each component attempts to maximise its own benefit \cite{goodfellow2014generative}. The generator receives a noise input sampled from a random distribution $z \sim p_{z}$ and learns to generate an output in the distribution $x \sim p_g$ that matches the structure of the unseen real data $x \sim p_{data}$. Meanwhile, the discriminator has access to the samples produced by the generator and the real data, and learns to distinguish between its real and fake inputs. While the output generated by $G$ improves during training as a result of the loss it obtains from the discriminator, the discriminator also becomes increasingly clever in recognising the data produced by the generator. Subsequently, GANs are particularly challenging to train since it must be guaranteed that both the generator and the discriminator maintain their competitiveness without outperforming each other early in the training phase. In the classic GANs, the generator and the discriminator attempt to maximise their objective by minimising the Jenson-Shannon Divergence (JSD), however, using JSD does not guarantee the convergence of losses, hence leading to training instability \cite{goodfellow2014generative}.

The Wasserstein GAN (WGAN) has been proposed as an alternative to the standard GANs in order to augment the training stability in generative models by replacing JSD with the Wasserstein Distance (WD) \cite{arjovsky2017wasserstein}. The use of the Wasserstein Distance ensures that the model is continuously learning even if the quality of the output is poor, and this is attributed to the smooth gradients produced by the Wasserstein cost function. The initial WGAN relied on weight clipping to enforce the confinement of the discriminator's weights within a specified range $\left[-c, c\right]$, however, the authors demonstrate that clipping can lead to difficulties with model optimisation \cite{arjovsky2017wasserstein}. Instead, the use of gradient penalty with WGAN has been proposed to mitigate against the exploding and vanishing gradients of the weights \cite{gulrajani2017improved}. In this setting, gradients are found using the linear interpolations $\hat{x} \sim p_{\hat{x}}$ between the real and the fake samples, where the distribution of linear interpolation is resembled by $p_{\hat{x}}$. Additionally, the gradient penalty coefficient $\lambda_{GP}$ is used as a parameter for controlling the level at which the gradient penalty affects the discriminator. 

The objective function for the WGAN-GP can therefore be represented as:

\begin{equation}
    \begin{aligned}
        \underset{G}{min} \: \underset{D}{max} \: V(D) 
        & \: \mathbb{E}_{x \sim p_{data}} \left[ D(x) \right]  -
         \mathbb{E}_{z \sim p_{z}}    \left[ D \left( G(z) \right) \right] \\
        & - \lambda_{GP} \: \mathbb{E}_{\widehat{x}
                \sim p_{\widehat{x}}}
                \left[
                    \left(
                        \Vert
                            \nabla_{\widehat{x}} D \left( \widehat{x} \right)
                        \Vert_{2}
                        -
                        1
                    \right)^{2}
                \right]
    \end{aligned}
\end{equation}

WGAN-GP is increasingly becoming more prevalent than the classic GANs in applications such as image generation and tabular data generation, as it contributes to more stable learning. In addition, WGAN-GP minimises the effect of mode collapse - that is when the generator learns to "trick" the discriminator by producing a limited number of modes which the discriminator incorrectly classifies as real samples, instead of utilising the entire data feature space. 

\subsection{Privacy Attacks}

In machine learning, membership inference attacks (MIA) aim to identify whether a data sample was used in the training of a machine learning model \cite{shokri2017membership}. For instance, the attackers might try to identify whether the records belonging a client were used for training a loan default prediction model. In this case, the attackers' objective would be to determine whether the client has taken a bank loan, with such information being used for targeted fraud attempts. Privacy guarantees in machine learning have been extensively studied in the form of analysing the connection to model overfitting \cite{yeom2018privacy} and in differential privacy \cite{dwork2008differential}. 

More recently, MIA have also been explored for generative models. Privacy attacks applied on synthetic samples generated by GAN models aim to reconstruct the real data samples which were used in GAN training. In principle, membership attacks on GANs can be categorised into three types of attacks \cite{hayes2019logan}; \emph{full black-box} in which attackers have access to only synthetic samples, \emph{partial black-box} where the attackers have access to the synthetic samples and the latent codes used to generate them, and \emph{white-box} which assumes that the attackers are able to access the internal parameters of the generator, the discriminator or both. The trade-off between the quality of synthetic samples and the privacy guarantees of GANs have been additionally examined in existing works \cite{xie2018differentially,jordon2018pate, chen2021gan}.


\section{Related Works}
\label{sc:related_works}

Tabular data is broadly used in regression and classification tasks, which facilitates a growing interest in tabular data synthesis for machine learning applications, especially in domains with limited training data. Bayesian networks can be used for generating synthetic records by approximating the conditional probability distribution from the data \cite{zhang2017privbayes}. While the Bayesian networks can in practice be additionally used for exploring causal relationships between the independent variables, estimating the distributions is often built on simplifying assumptions on the data \cite{goncalves2020generation}. Meanwhile, tree-based methods were first utilised for generating partial synthetic data in \cite{reiter2005using}, and has further explored in \cite{watson2023adversarial} where adversarial random forest has demonstrated comparable performance to deep learning techniques in terms of synthetic data quality. However, the privacy-utility trade-offs for synthetic data generation using tree-based density estimators is not sufficiently explored.

Deep neural networks have been widely studied for synthetic data generation, thanks to their capabilities for handling and approximating the distributions of large datasets. Variational auto-encoders (VAEs) \cite{kingma2013auto} estimate the probabilistic distribution of by finding a lower-dimensional latent representation of the data. The application of VAEs has been extended to image data generation \cite{wan2017variational}, oversampling of anomaly event data \cite{islam2021crash} and tabular data synthesis \cite{xuModelingTabularData2019}. An underlying limitation with variational autoencoders is the assumption of the a simple parametric form of the latent space, which leads to a difficulty in capturing complex data distributions \cite{higgins2017beta}. Invertible neural networks have also been proposed for tabular data synthesis through variants based on neural ordinary differential equations \cite{lee2021invertible}, copula flows \cite{kamthe2021copula} and normalizing flows for private tabular data generation \cite{lee2022differentially}. A drawback of the invertible neural network based synthesis is that extensive hyperparameter tuning is needed to achieve satisfactory classification and regression performance.

\begin{table*}[!htb]
\caption{Properties of some state-of-the-art tabular GANs, used as baselines later Section \ref{sc:results} along with our proposed model, CasTGAN. When counting the number of datasets, we only consider real datasets as opposed to simulated datasets with known distributions and priors. While some methods conduct privacy analysis by measuring distance to closest records, this is notably different than evaluating the threat posed by privacy attacks.}
\label{tb:baseline_overview}
\small
\centering
\resizebox{0.98\textwidth}{!}{
    \begin{tabular}{l l l l l l l l}
        \toprule
        Method & Data types & No. of real datasets & Machine learning utility evaluation & Univariate similarity evaluation & Correlation analysis & Validity evaluation & Privacy attacks evaluation \\
        \midrule
        table-GAN \cite{parkDataSynthesisBased2018a}  & mixed & 4 & \cmark & \cmark & \xmark & \xmark & \cmark  \\
        medGAN \cite{choiGeneratingMultilabelDiscrete2017}  & continuous & 3 & \cmark & \cmark & \xmark & \xmark & \cmark  \\
        CTGAN \cite{xuModelingTabularData2019}  & mixed & 6 & \cmark & \cmark & \xmark & \xmark & \xmark  \\
        cWGAN \cite{engelmannConditionalWassersteinGANbased2021}  & mixed & 7 & \cmark & \cmark & \xmark & \xmark & \xmark  \\
        CTAB-GAN \cite{ctab2021zhao} & mixed & 5 & \cmark & \cmark & \cmark & \xmark & \xmark  \\
        CasTGAN [ours] & mixed & 6 & \cmark & \cmark & \cmark & \cmark & \cmark  \\
        \bottomrule
    \end{tabular}
}
\end{table*}

Within deep learning based generative models, GANs are favourable due to their ability to generate complex synthetic data in an adversarial and unsupervised setting \cite{pan2019recent}. table-GAN \cite{parkDataSynthesisBased2018a} is one of the earliest generative models for producing tabular synthetic output based on adversarial training. Using convolutional neural networks, table-GAN demonstrates that GANs unsurprisingly outperform anonymisation techniques while highlighting the potential privacy risks arising from membership attacks. Meanwhile, in medGAN \cite{choiGeneratingMultilabelDiscrete2017} an autoencoder based generative model is developed for generating high-dimensional medical patient records while shedding light on the privacy risks attributed to the generated data. A long-short term memory (LSTM) architecture for the generator was adopted in \cite{xu2018synthesizing} demonstrating the potential of recurrent neural networks in synthetic data generation. A GAN approach based on an autoencoder and Kullback-Liebler divergence to tackle mode-collapse was proposed in  \cite{srivastava2017veegan}, in which high-quality synthetic output was produced, with comparable performance to state-of-the-art methods. In CTGAN \cite{xuModelingTabularData2019}, conditional training of a GAN is carried out by instructing the generator and the discriminator to sample based on randomly selecting feature category in every training iteration, where a highly realistic tabular output can be observed from their evaluation. In \cite{engelmannConditionalWassersteinGANbased2021}, the authors propose cWGAN, which is a GAN-based oversampling technique focusing on the generation of samples belonging to the minority class in financial credit datasets. Zhao et. al \cite{ctab2021zhao} builds up on existing tabular GAN models by employing convolutional neural network (CNN) and conditional vectors to improve the representation of skewed distribution of numerical features of the synthetic output. Finally, Strelcenia et al. \cite{strelcenia2023survey} comprehensively review the existing tabular GAN literature and highlight that an underlying limitation of the tabular GAN-based studies is the lack of standarised evaluation metrics.

Acknowledging the aforementioned, it is manifested that there is no shortage of novelties in synthetic tabular data generation literature. Nevertheless, an underlying challenge remains the proposal of evaluation techniques and criteria for quantifying the reliability and the statistical properties of the synthetic data. A further limitation is the sufficient analysis of data with hierarchical and interdependent variables. Therefore, our focus in this work is proposing a new framework that alleviates the two preceding deficiencies in the synthetic tabular data domain. A general overview of the properties, strengths and limitations of our model and the existing studies that we adopt as baselines is outlined in Table \ref{tb:baseline_overview}.

\section{Methodology}
\label{sc:methodology}

Generating synthetic data from unknown and correlated distributions is a non-trivial task. The architecture of CasTGAN is tailored for generating mixed-type features that have similar distributions to the ones observed in a real dataset. Additionally, cascaded structures enable modelling correlations among features in a sequential manner. Given a dataset $X$ with $M$ features, the features of the dataset can be represented as $\{m_1, m_2, \ldots, m_M\}$. 

\subsection{Model Architecture}

\begin{figure*}[!htbp]
    \centering
    \includegraphics[width=0.9\textwidth]{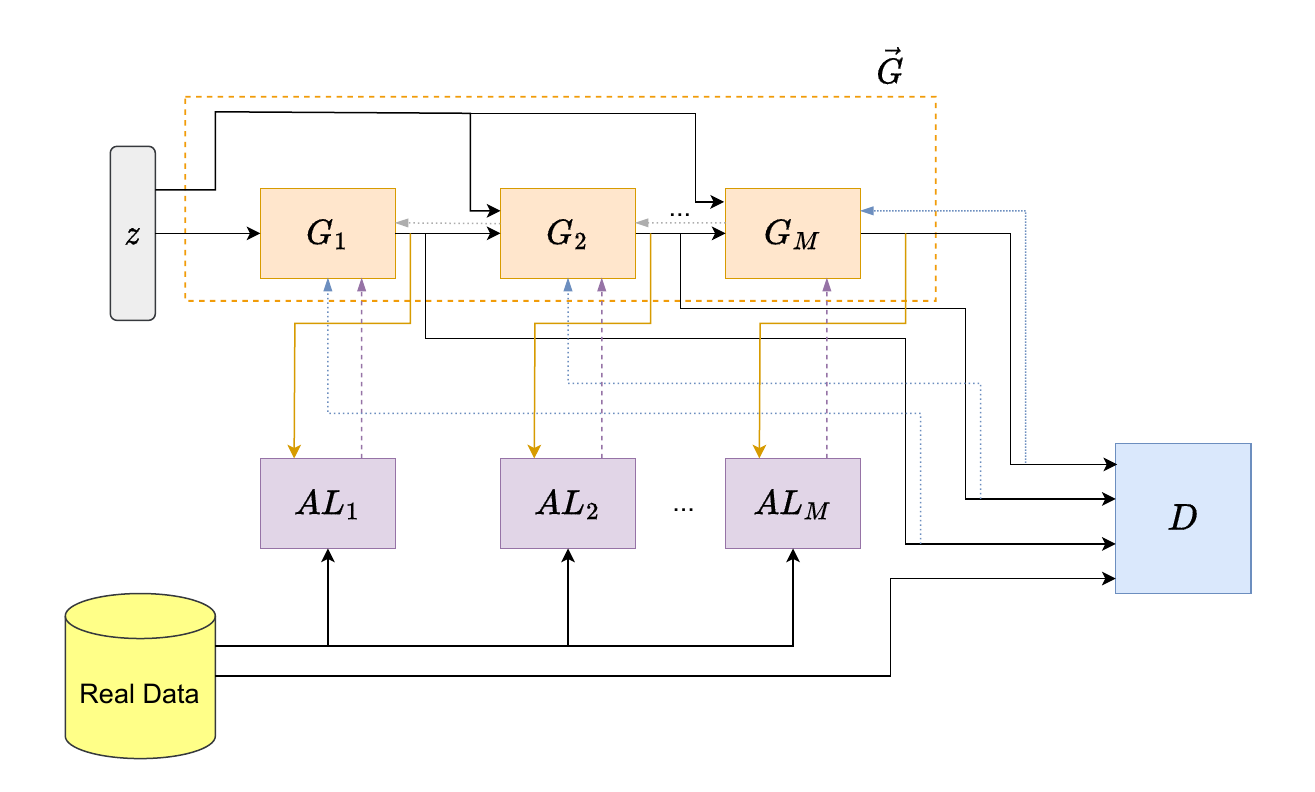}
    \caption{The model architecture of CasTGAN. The cascade $\vec{G}$ is composed of generators $G_{1}, G_{2}, \ldots, G_{M}$ sequentially lined up. The auxiliary learners $AL_{1}, AL_{2}, \ldots, AL_{M}$ are fitted on the real data, and are utilised by their respective generators for querying the generation of the data features $m_{1}, m_{2}, \ldots, m_{M}$. The cascade of generators $\vec{G}$ takes noise vector $z$ as input, while the discriminator $D$ is trained to distinguish the real data from the synthetic data. As depicted, generators $G_{1}, G_{2}, \ldots, G_{M-1}$ receive three losses: loss directly from the discriminator, loss backpropagated from the previous generator and the loss from the auxiliary learner. Meanwhile, $G_{M}$ is passed the loss from its auxiliary learner and the loss from the discriminator.}
    \label{fig:ganstruct}
\end{figure*}


The proposed CasTGAN framework is characterised by a cascade $\vec{G}$ of multiple generators, $\vec{G}=\{ G_{1}, G_{2}, \ldots, G_{M}\}$, in which generators $G_{1}, G_{2}, \ldots, G_{M}$ are connected sequentially and coupled with auxiliary learners $AL_{1}, AL_{2}, \ldots, AL_{M}$. Generator $G_{i}$ and auxiliary learner $AL_{i}$ for $i = 1, 2,\ldots, M$ are devoted for feature $m_i$ in the dataset, and the real data is used for fitting the auxiliary learners and the discriminator $D$. An illustration of the CasTGAN architecture is depicted in Figure \ref{fig:ganstruct}.


As can be visualised from Figure \ref{fig:ganstruct}, each generator $G_{i}$ focuses on generating its target feature using a primary neural network. The cascade of generators are laid out sequentially such that generator $G_{i}$ obtains its inputs from a given noise vector $z$ whose components are standard Gaussian and i.i.d., and from the outputs of the previous generator -- the only exception being the first generator which only takes a vector of random noise as its input.

\begin{figure*}[!htbp]
    \centering
    \includegraphics[width=0.95\textwidth]{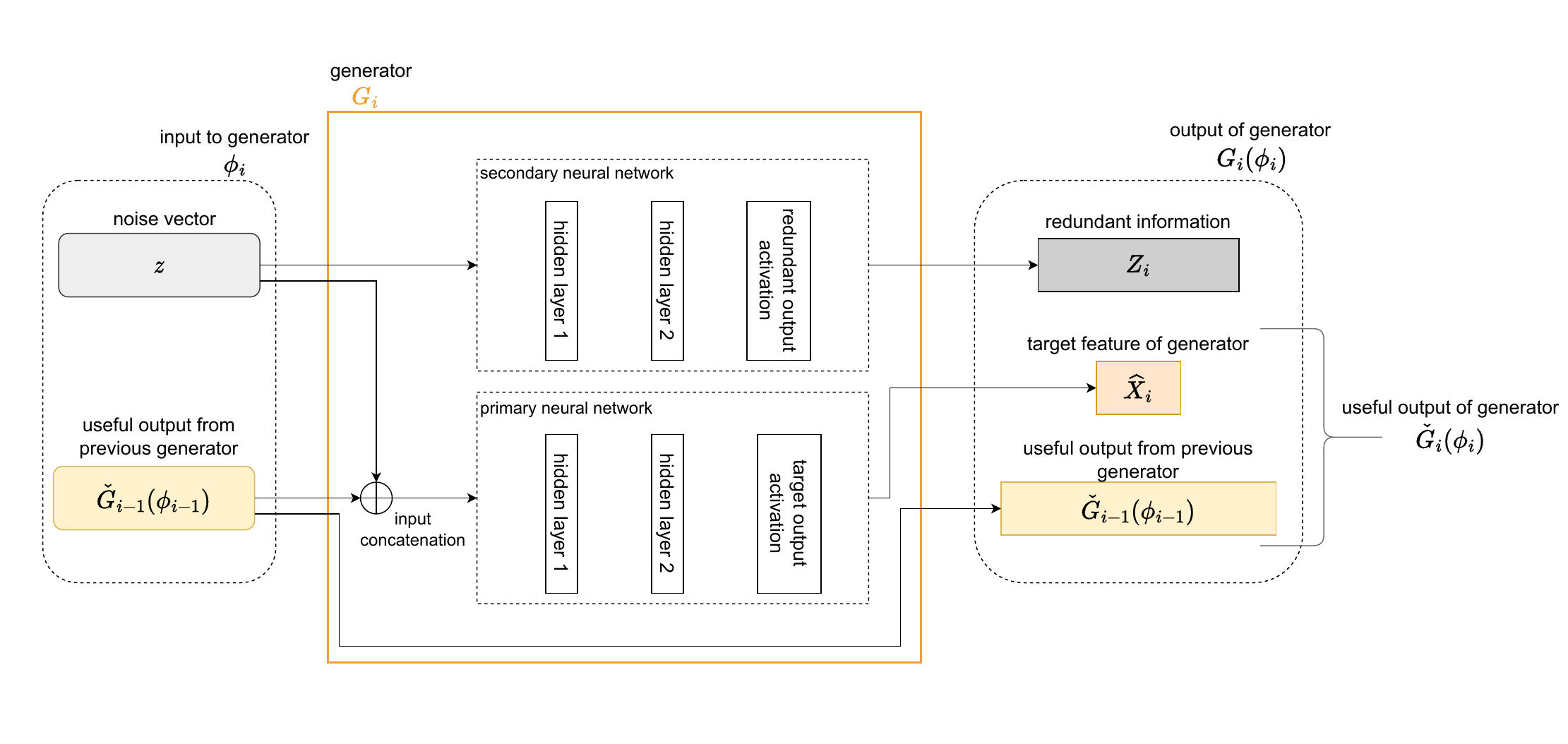}
    \caption{A close in visualization of generator $G_{i}$ in our GAN architecture. The structure in the figure is applicable to all generators in the cascaded layout, except for $G_{1}$, which receives only noise vector $z$ as input.}
    \label{fig:glogic}
\end{figure*}


Notation-wise, generator $G_{i}$ takes as input $\phi_{i}$ two objects: the useful outputs coming from $G_{i-1}$ (i.e., the vector $\check{G}_{i-1}(\phi_{i-1})$) and the noise vector $z$ (note that the same vector is fed to all the generators as depicted in Figure~\ref{fig:ganstruct}). The generator $G_{i}$ then produces one output, i.e., $G_{i}(\phi_{i})$, that may though be logically split in three distinct components: $Z_{i}$, that will be considered redundant information and that will not be used by the next generator; $\widehat{X}_{i}$, that is the target feature of generator $i$; and $\check{G}_{i-1}(\phi_{i-1})$, that is simply the information from the past generator that will be forwarded to the next one. The input and the output of generator $G_i$ are shown in Figure~\ref{fig:glogic}.

Formally, the output of generator $G_{i}$ can be presented as:
\begin{equation}
    G_{i}(\phi_{i}) = \check{G}_{i}(\phi_{i}) \oplus Z_{i} ,
\end{equation}
while the information that generator $G_{i}$ will pass to $G_{i+1}$ is
\begin{equation}
    \check{G}_{i}(\phi_{i})
    = 
    \begin{cases}
        \widehat{X}_{i} & \text{if } i = 1 \\
        \widehat{X}_{i} \oplus \check{G}_{i-1}(\phi_{i-1})   & \text{if } i \geq 2 \; .
    \end{cases}
\end{equation}
Note that the generator is actually composed by two distinct neural networks: the primary one, whose input is
\begin{equation}
    \phi_{i} = 
    \begin{cases}
    z & \text{if } i = 1 \\
    z \oplus \check{G}_{i-1}(\phi_{i-1}) & \text{if } i \geq 2 ,
    \end{cases}
\end{equation}
and the secondary neural network, whose input is the noise vector $z$ above and whose output $Z_{i}$ is the redundant information output mentioned above, that will not be passed forward to $G_{i+1}$ but will instead be used by $AL_{i}$.

We note that, as the losses are not backpropagated to the secondary neural network, $G_{i}$ retains its primary objective of generating its target feature based on the input provided to it. Summarizing, the overall cascaded generator structure can be denoted as
\begin{equation}
    \vec{G} \left( z \right)
    =
    \widehat{X} ,
\end{equation}
where $\widehat{X}$ is the generated synthetic output.

Based on our literature survey, we observe that some of state-of-the-art tabular GANs employ a conditional setting to enforce the representation of features in both the generator and discriminator \cite{xuModelingTabularData2019, ctab2021zhao}. We note that while this is indeed an effective strategy for representing discrete categories and preventing mode collapse, this approach is considerably inefficient for sampling datasets with a large number of categories and few data records, since conditioning on a single random category at every training iteration might not be sufficient to cover all the existing categories in the dataset. In this paper we seek to analyse whether, how much and under which conditions resorting to a series of auxiliary learners -- one for each feature -- may encourage the models to learn to represent based on the losses traversed, rather than explicitly constraining the model output. The hypothesis is indeed that if multiple auxiliary losses are computed in parallel, the model might be able to improve the learning of categorical interdependence and scale up accordingly to highly dimensional tabular datasets.

\subsection{Auxiliary Learners Architecture}

GANs for tabular data synthesis are known to be prone to training instability and mode collapse due to the imbalanced feature categories \cite{kim2021oct}. Conditional GANs \cite{mirza2014conditional} have been deployed to generate synthetic output belonging to specific classes. Conditioning both on the generator and the discriminator has been shown to stabilise the training process of GANs.

On the other hand, the use of auxiliary learning for predicting the target variable given the data represents an alternative approach for capturing the characteristics of the data attributed to given target feature. It has been demonstrated that the auxiliary loss further stabilises the training process in comparison to conditional generation, and leads to a representation that is independent of target label \cite{waheed2020covidgan}. We observe that while auxiliary learners are traditionally embedded within the discriminator \cite{odena2017conditional}, we instead propose designing auxiliary learners as independent structures. 

In the CasTGAN, we craft M auxiliary learners $AL_{1}, \ldots, AL_{M}$ for learning to predict the individual features. Due to its scalability on large datasets and the relatively fast convergence speed, we focus on building the auxiliary learners using the Light Gradient Boosting Machine (LightGBM) \cite{ke2017lightgbm}, which we pre-train prior to the GAN training. An auxiliary learner $AL_i$ corresponding to feature $m_i$ is trained on $X_{\not\in i}$ in order to predict $X_i$. Following standard strategies for such tasks, for predicting the numerical features, the mean-squared error loss is used in the training of the auxiliary learners, whereas cross-entropy loss is used for predicting the categorical and binary variables. As with other decision tree based models, there is no need to one-hot encode the categorical features in $X_{\not\in i}$, but instead the categories are converted into integer encodings. Meanwhile the LightGBM auxiliary learners are capable of handling numerical features with extreme magnitudes, and therefore numerical features are not scaled for auxiliary training.

As LightGBM models have low computational complexity and are generally fast to to train \cite{ke2017lightgbm}, assigning an auxiliary learner for every feature is a reasonable approach for representing the auxiliary loss $L_{AL}$ for predicting a feature given all the other features. It is worth noting that for the early auxiliary learners in the cascaded sequence $AL_1 \text{ to } AL_{\lceil M/2 \rceil}$, the generated data feature space $\widehat{X}_{\not\in i}$ is heavily dominated by redundant variables $Z$ which subsequently lead to increased auxiliary losses. However, these losses help the early generators in producing features that closely match the distributions of the training data. Meanwhile, the task for the later generators and auxiliary learners in the cascade becomes increasingly focused towards generating features that can be predicted from the initially generated target features $\check{G}_{i-1}\left(\phi_{i-1}\right)$.

\begin{figure}[!htbp]
    \centering
    \includegraphics[width=\columnwidth]{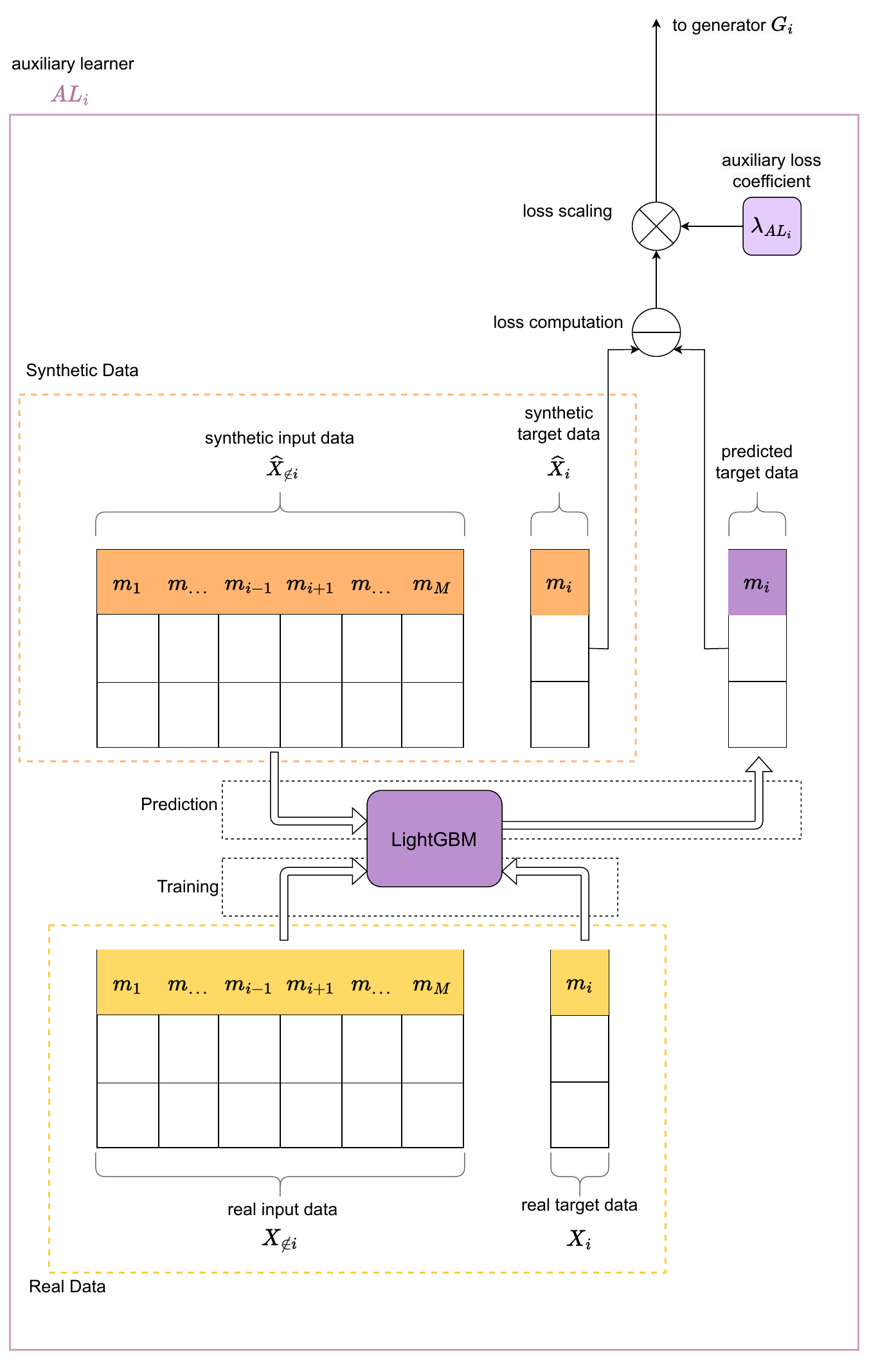}
    \caption{A detailed depiction of auxiliary learner $AL_{i}$ in CasTGAN. The training of the LightGBM model on the real data occurs prior to the training of the GAN model. Meanwhile, the synthetic data from the generator $G_{i}$ is queried against the auxiliary learner $AL_{i}$ during the training iterations of the GAN to compute the auxiliary loss of the generator's target feature.}
    \label{fig:aux_logic}
\end{figure}

To ensure that the auxiliary losses $L_{AL_1}, \ldots, L_{AL_M}$ do not overexceed the generator's ones, there is the need for scaling down the losses from the auxiliary learners. In this paper we analyze the choice of performing this scaling down by means of constant coefficients $\lambda_{AL}$. In principle, $\lambda_{AL}$ could be a single scalar value that is applied to all the auxiliary losses. However, we consider a vector of auxiliary loss coefficients $\lambda_{AL_1},  \ldots , \lambda_{AL_M}$ since it is fundamentally important for the early generators to generate variables that conform to the original feature distributions, since this will prompt the next generators in the cascade to effectively learn the feature correlations. Though this a hyperparameter, we set $\lambda_{AL_1} = 0.75$ and $\lambda_{AL_M} = 0.10$ while the auxiliary coefficients in between are linearly and equidistantly scaled in the $[0.75, 0.10]$ range in our experiments. The same auxiliary setting is applied for all the datasets in this work. The overall structure of auxiliary learner $AL_i$ is illustrated in Figure~\ref{fig:aux_logic}.

\subsection{Training Data Transformation}

The main novelty in this paper comes from testing the effects of designing multiple generators in a cascaded layout, where each generator focuses on generating a single data feature. To represent the numerical features we propose to use Variational Gaussian Mixture models (VGM) \cite{blei2006variational} to estimate the number of modes for a numerical feature, as it has been demonstrated that correctly representing multi-modal numerical data objectively reduces the incidence of mode collapse \cite{srivastava2017veegan}. In this context, each mode is essentially a Gaussian model on its own, where in the transformation process a single mode is selected for a feature and a scalar value is calculated for quantifying the magnitude of the mode. As such, the transformation of a numerical feature changes the initial unscaled real number into a vector of size equivalent to the number of Gaussian mixture models + 1 (the \emph{1} being the magnitude value of the respective mode and the vector being a one-hot encoded representation of the selected mode). The representation of continuous features using variational Gaussian models is not exclusive to this work as it has been adopted with notable success in earlier tabular GANs \cite{xuModelingTabularData2019,ctab2021zhao}.

Meanwhile, the categorical features of the training data are transformed into one-hot encodings before being fed to the discriminator. For GANs, the one-hot encoding vectorisation of the categorical features presents an intuitive approach to process the data as it can be scaled and can be appropriately used by the model without issues such as exploding gradients. Furthermore, the use of one-hot encoding simplifies the task of introducing non-linearities by the generator for guaranteeing that the model gradients are differentiable. It is worth reiterating that categorical and numerical transformations of the data for use by the generator and discriminator differ from the representation of the same data used for training and evaluating the auxiliary learners.

\subsection{Generators and Discriminator}

The generators receive input in the form of noise vector $z$ and the untransformed meaningful output $\check{G}_{i-1}(\phi_{i-1})$. As with other GAN applications, we highlight that using a larger noise vector leads to a better output of the features and can mitigate against mode collapse \cite{srivastava2017veegan}. Throughout all the experiments, we use a noise vector of size 128, though this is a hyperparameter that can be tuned accordingly \cite{salimans2016improved}. Since each generator dedicates its effort into generating one feature at a time, we use a simple primary neural network of hidden sizes $(128, 64)$. Additionally, we use layer normalisation after the hidden dimension~\cite{ba2016layer} for standardizing the weights into zero mean and unit variance and for speeding up the training process. We also use the LeakyReLU activation function with a small negative slope as opposed to ReLU in order to remove the constraints associated with setting the negative gradients to zero. 

The dimension of the output layer of the generator for producing the target feature is equivalent to the number of one-hot encodings if the feature is categorical or equal to the number of VGM modes $+1$ if the target feature is numerical. A hyperbolic tangent (\texttt{tanh}) activation is applied to the scalar value of the numerical VGM representation. For the categorical output and the one-hot encoded vector of the VGM vector we use gumbel softmax activations for introducing non-linearities to the output. The Gumbel-softmax works by adding noise from the Gumbel distribution to the vectorised logit output of the generator while maintaining the differentiable nature of the GAN training. The Gumbel-softmax function exhibits also a temperature parameter $\tau$ that may be used to control the diversity of the output generated by the function. In our experiments $\tau = 0.8$ was assessed as proper to generate a diversified output that reduces the effects of mode collapse, while conforming to the distribution of categories of the feature within the training set.

We then note the risk that the discriminator may learn to distinguish between real and generated data by discriminating between the hard one-hot encoded real data and the float values from the generator. To minimize this risk we add an i.i.d. Gaussian noise distributed as $\mathcal{N}(0,0.01)$ \cite{arjovsky2017wasserstein} to the columns of the real samples before feeding them to the discriminator. Consequently, all the inputs to the discriminator (i.e., numerical and categorical features of real and generated data) are float values. The weights of the discriminator are then trained using only the outputs from the final generator $G_M$. The parameters of the various generators are though updated based on the loss of the discriminator, that is thus computed for this reason. 

We maintain a simple architecture for the discriminator comprising two hidden layers of sizes $(256, 128)$. As with the generator, layer normalization and LeakyReLU activations are used between the hidden layers. The final layer consists of a single output node without an activation function. To alleviate against mode collapse \& GAN training instability issues we additionally compute the Wasserstein loss~\cite{arjovsky2017wasserstein} with gradient-penalty~\cite{gulrajani2017improved} for the calculation of the discriminator losses.

As the CasTGAN is built up using $M$ generators, we have multiple min-max games between the generators and the discriminator. Therefore, the value function for the discriminator can be expressed as
\begin{equation}
    \begin{aligned}
        \underset{\vec{G}}{\min} \: \underset{D}{\max} \: V(D)
        =
        & \phantom{-}  \: \mathbb{E}_{x \sim p_{data}} \left[ D(x) \right] \\
        & - \: \mathbb{E}_{z \sim p_{z}}    \left[ D \left( \vec{G}(z) \right) \right] \\
        & - \: \lambda_{GP} \mathbb{E}_{\widehat{x}
                \sim p_{\widehat{x}}}
                \left[
                    \left(
                        \Vert
                            \nabla_{\widehat{x}} D \left( \widehat{x} \right)
                        \Vert_{2}
                        -
                        1
                    \right)^{2}
                \right]
    \end{aligned}
\end{equation}
and value function for generator $G_i$ is hence given by
\begin{equation}
    \begin{aligned}
        \underset{G_i}{\min} \: \underset{D}{\max} \: V(G_i)
        =
        & - \: \mathbb{E}_{\phi \sim p_{\phi}}
                \left[ D \big( G_{i}(\phi) \big) \right] \\
        & + \: \lambda_{AL_i} \: \mathbb{E}_{\phi \sim p_{\phi}}
                \left[ L_{AL_i} \right] .
    \end{aligned}
\end{equation}

\section{Experimental Setup}
\label{sc:exp_setup}

Evaluating the performance of GANs is a non-trivial task, and this is evident from literature, where no standard approach for evaluating GANs on tabular data can be found. In this section, we describe the experimental design for CasTGAN, with a thorough discussion of the metrics used for the model's analysis. We implemented our model using PyTorch in Python on a Linux Ubuntu 20.04 machine running on AMD Ryzen Threadripper 3990X and Nvidia GeForce RTX 3090. Our CasTGAN source code is publicly available\footnote{https://github.com/abedshantti/CasTGAN}. 

\subsection{Datasets}

Our CasTGAN is designed for synthesis of tabular data that can be typically found in the financial and healthcare sectors. We therefore use tabular mixed datasets characterised by a combination of categorical and numerical features. We use four datasets where the task is the binary classification of the target label - Adult \cite{data_adult}, Bank Marketing \cite{data_banking}, Taiwan Credit \cite{data_taiwan} and Diabetes \cite{data_diabetes}. Meanwhile, we use the House Prices \cite{data_housing} and Cars \cite{data_cars} datasets for regression. We additionally highlight that binary columns in the datasets are handled as categorical variables. An overview of the datasets used is presented in Table \ref{tb:datasets}. For synthesising data with CasTGAN and other baselines, we use 50\% of the datasets' total number of samples for training the models. The remaining 50\% of the data samples is dedicated for evaluating the generated synthetic output.

\begin{table}[!htb]
\caption{Datasets used in this study.}
\label{tb:datasets}
\small
\centering
\resizebox{\columnwidth}{!}{
    \begin{tabular}{l l l l l l}
        \toprule
        Dataset &  Samples & Num. & Cat. & Unique & Task \\
        & & features & features & categories & \\
        \midrule
        Adult & 32561 & 6 & 9 & 104 & Binary classification \\
        
        Bank  & 45211 & 10 & 11 & 55 & Binary classification \\
        
        Credit  & 30000 & 14 & 10 & 79 & Binary classification \\
        
        Diabetes  & 253680 & 7 & 15 & 30 & Binary classification \\
        
        Cars & 472336 & 5 & 8 & 2409 & Regression \\
        
        Housing  & 21613 & 17 & 2 & 72 & Regression \\
        
        \bottomrule
    \end{tabular}
}

\end{table}

\subsection{Baselines}

We compare the synthetic output of CasTGAN against five state-of-the-art tabular generative adversarial network models: table-GAN \cite{parkDataSynthesisBased2018a}, medGAN \cite{choiGeneratingMultilabelDiscrete2017}, CTGAN \cite{xu2018synthesizing}, cWGAN \cite{engelmannConditionalWassersteinGANbased2021} and CTAB-GAN \cite{ctab2021zhao}. Given that some of the datasets that we use in this study were not evaluated previously by the existing methods, we selected the optimal hyperparameters recommended by the baseline methods' authors in this study.

\subsection{Hyperparameter selection}

We emphasise that while our framework comprises an ample number of hyperparameters, we state that our reasoning for refraining from fine-tuning our model's hyperparameters in our experimental results is twofold. First, an extensive hyperparameter tuning would need to be conducted on a data-level basis and on a criteria level. To demonstrate that our framework is compatible with any dataset, we conduct our experimental analysis and train CasTGAN on all the datasets in this work using the same set of parameters in Appendix~\ref{app:def_hyper}. The choice of these parameters is based on preliminary, yet limited experimentation of our framework which yielded satisfactory and promising synthetic output during the conceptualisation phase. From a criteria-level perspective, we highlight that the model settings that attain the best performance on an evaluation criterion do not necessarily improve the synthetic output on all performance evaluation fronts. For instance, there might exist a hyperparameter trade-off between maximising the machine learning utility and improving the univariate feature representation. Second, given that we refrain from conducting hyperparameter tuning for the benchmark models and instead adopt the recommended hyperparameters of the benchmarks across all datasets, the comparison between our method and the baselines can be regarded as fair and equitable, provided that no extensive tuning of CasTGAN is undertaken correspondingly. Notwithstanding, we refer the reader interested in strategies for selecting hyperparameters for training CasTGAN on their own data to Appendix~\ref{app:hyper_guide}.

\subsection{Evaluation Criteria}

\subsubsection{Train on Synthetic, Test on Real (TSTR)}

We measure reliability of the synthetic output produced by CasTGAN by training machine learning models on the generated data. We fitted three machine learning models on the generated output - namely AdaBoost, random forest and logistic regression for classification tasks and AdaBoost, decision trees and Linear SVM for regression tasks. We then used the trained models to predict the target label of the test data and we report the precision-recall area under curve (PR-AUC) for binary classification tasks and the $R^2$ score for regression tasks. Since our main objective is to measure the machine learning utility of the generated data rather than assessing the individual performance of each classifier, we average the metrics produced by the three machine learning models.

\subsubsection{Univariate Distributions}

We also assess the extent at which the individual features generated by CasTGAN resemble the features of the real data. As we quantitatively analyse how well our model learns the univariate feature distributions, we first one-hot encode and normalise the synthetic and real datasets. We calculate the dimension-wise mean of the individual features of the synthetic output and training data and report the RMSE between the real and the synthetic output. Additionally, we report the Kolmogorov-Smirnov two-sample test score \cite{chakravarti1967handbook} between the real univariate features and the synthetic ones.

\subsubsection{Correlation and Diversity}

Measuring the validity of the synthetic output represents a significant challenge for GAN frameworks. In computer vision applications of GANs, the synthetic images can be in some cases distinguished from the real images by a human observation of irregularities in the output such as pupil orientation in human eyes and out-of-place pixels. In tabular GANs, similar challenges exist as there is no standard approach in the literature for quantifying the proportion of invalid samples in the synthetic output. For instance, a common example in the tabular synthesis literature is highlighting how an entry such as gender = "Female" and diagnosis = "Prostate Cancer" is by definition an invalid record, as no such entry exists in the original data, nor can a female be diagnosed with prostate cancer by a physician. In financial datasets, such nested relationships also exist between features if for example we inspect a dataset with a city column and a country column. In such cases, a record with city = "Buenos Aires" and country = "Malta" is by definition invalid. Meanwhile, entries with city = "Alexandria" and country = "United States" is valid as Alexandria exists in the United States even though it is more commonly attributed to the country "Egypt". It is for this reason that quantifying the invalid output generated by tabular GANs is no easy task, even in the presence of domain knowledge.

While a GAN model needs to ensure that its synthetic output is as valid as possible, there also needs to be some considerations for the diversity of the generated output. As such, the generative models should not significantly restrict the possible feature combinations between the different categorical features. Ensuring the diversity of the synthetic output enables the model to be less deterministic and increases its robustness against privacy attacks that aim to identify sensitive information. Therefore, the GAN model should be encouraged to explore unique feature combinations as long as such combinations can be considered valid.

Given that public datasets are used in this work for the purpose of reproducability, we do not have the full domain knowledge for these datasets, thus, we propose an alternative method for quantifying the validity of the synthetic data. First, we consider calculating the difference in feature correlations between the training data and the fake data. For computing the correlations between numerical features we use the Pearson's correlation coefficient, while the Cramer's V measure is used for capturing the correlation between categorical features. The correlation score is found by calculating the root mean squared error (RMSE) score between the elements of the triangular matrix of the synthetic dataset and the real dataset.

For measuring the diversity of the categorical output, we propose a new metric - Unique Pairwise Categorical Combinations (UPCC). In essence, we count the total number of unique interactions between any pair of categorical features in the dataset. For instance, in the Adults dataset the combinations [education = "Bachelors" and marital-status = "Never-married"], [education = "Bachelors" and sex = "Male"] and [marital-status = "Never-married" and sex = "Male"] each counts as a single pairwise combination, regardless of how many times they appear in the data. A reliable model therefore ensures that the UPCC of its output should be comparable to that of the original data. Subsequently, the UPCC Ratio is the number of unique pairwise combinations of synthetic output divided by the total number of unique combinations of the training data. 

Finally, we estimate the validity of the model's output by dividing the correlation RMSE score of the model by the UPCC Ratio. We name this measure as the CORDV score. A lower CORDV score indicates that the model is able to minimise the difference in feature correlation between its synthetic output and the real data, while simultaneously not impeding its ability in generating unique feature combinations. Meanwhile, a worse generative model is reflected by a greater CORDV score, indicating that the model poorly captures the correlations while potentially restricting the uniqueness of the categorical pairs. 

\subsubsection{White-Box Privacy Attacks}

Traditionally, white-box membership inference attacks on GANs assume that the attacker has access to the synthetic data and at least one generative component of the model. In this work, we formulate white-box privacy attacks in a different setting. We highlight that while using multiple auxiliary learners help in generating more realistic and reliable synthetic output, the use of multiple auxiliary learners leads to a more susceptible model for privacy breaches by attackers. 

In this work, we devise white-box attacks by assuming that an attacker has access to the trained auxiliary learners and attempts to reconstruct training samples through an iterative process of estimating a hidden feature. In essence, the attacker with the synthetic data will at a given time remove one column from the data, use the corresponding the auxiliary learner to predict the masked feature using the remaining features, and then replacing the masked column with the predicted output from auxiliary learners. In this setting, a single iteration refers to a walk-through over all the auxiliary learners and subsequently replacing all the columns in the dataset once.

For evaluating how effective such white-box attacks on our model, we control the training the of the auxiliary learners using a perturbation parameter $\epsilon$. The perturbation parameter translates to the proportion of label samples that are modified when training the auxiliary learners prior to the GAN training. For an auxiliary learner corresponding to a numerical column $X_i$, we perturb the numerical variables such that perturbed variable for a given sample $\tilde{x}$ can be calculated as $\tilde{x}_i = x_i + \alpha x_i $, where $\alpha$ is a floating number randomly sampled from $[-1.0, 1.0]$. Meanwhile, we perturb the categorical features by randomly selecting a category from the list of all the unique categories of the said feature. In our analysis, we experiment with $\epsilon = 0.0$, implying that no perturbation takes place, and gradually increment this value to $\epsilon = 0.3$, implying that 30\% randomly chosen samples for each auxiliary learner were perturbed prior to the auxiliary training.

Furthermore, we analyse whether an attacker possessing the original data preprocessing transformers has an additional advantage in recovering the training samples. We hypothesise that an attacker with access to the data transformations used for the auxiliary training can simply convert the synthetic to a data structure that aligns with the existing transformer. On the other hand, an attacker without access to the preprocessors needs to fit and transform the data independently before launching membership attacks. This is especially prevalent for categorical features, where transforming the categories into ordinal encodes that do not match the ones learned by the auxiliary learners might lead to less effective attacks.


\section{Results}
\label{sc:results}

\subsection{Machine Learning Utility}

We use the synthetic data generated by CasTGAN for fitting machine learning classification and regression supervised models on the six datasets and compare our performance on the test set against models fitted on the training set and models fitted on the five synthetic output of the five baseline methods. Additionally, we compare the performance of our model and the other baselines against the real training datasets used for fitting the predictive models, which we refer to as \emph{Identity}. We note that we were unable to run the highly-dimensional Cars dataset on CTAB-GAN to exceedingly large memory requirements attributed to representing very high-dimensional data in the convolutional neural network GAN-based approach. The results are computed in Table~\ref{tb:ml_util}.

\begin{table}[!htb]
\caption{Binary classification (PR-AUC score) and regression ($R^2$ score) evaluation on the test sets.}
\label{tb:ml_util}
\centering
\resizebox{0.95\columnwidth}{!}{
    \begin{tabular}{l cccc|cc}
        \toprule
        &  Adult & Bank & Credit & Diabetes & Cars &  Housing\\
        \midrule
        Identity & 0.7744 & 0.6085 & 0.5374 & 0.3980 & 0.7604 & -0.3696 \\
        \hline
        table-GAN & 0.2225 & 0.0954 & 0.2084 & 0.1585 & -2.0148 & -83.1326 \\
        medGAN & 0.5777 & 0.2729 & 0.3050 & 0.3241 & \textbf{0.7491} & \textbf{-0.3845} \\
        CTGAN & 0.6932 & 0.4805 & 0.4558 & 0.3711 & 0.5327 & -0.6533 \\
        cWGAN & 0.3030 & 0.2692 & 0.2166 & 0.3325 & -319.0963 & -1.1054 \\
        CTAB-GAN & \textbf{0.7192} & 0.4920 & 0.4911 & 0.3589 & - & -0.7284 \\
        CasTGAN & 0.6718 & \textbf{0.5657} & \textbf{0.4995} & \textbf{0.3866} & 0.5566 & -0.4433 \\
        \bottomrule
        
    \end{tabular}
    }
\end{table}

From Table~\ref{tb:ml_util} we can observe that the TSTR metrics for our CasTGAN is consistently within the best performing synthetic output, outperforming all the baselines on three out of six datasets. On the Bank dataset, we can observe that CasTGAN's PR-AUC score of 0.5657 ranks closer to the classification models trained on the Bank real training data with a score of 0.6085, than the second best classification by CTAB-GAN with a score 0.4920. Similarly, CasTGAN falls 0.0114 short of the PR-AUC score exhibited by the real data on the Diabetes dataset, whilst outperforming the second best synthetic model, CTGAN, with a PR-AUC difference of 0.0155. We also observe that CasTGAN also ranks first among the GAN models on the Credit dataset, despite being closely challenged by CTAB-GAN. For the datasets which CasTGAN did not achieve the highest machine learning utility, the results in Table~\ref{tb:ml_util} show that CasTGAN narrowly underperformed against two of the baselines on the Adult dataset. Furthermore, we note that medGAN demonstrated the best results on regression datasets, whereas, synthetic data from CasTGAN followed as the second best in terms of the $R^2$ score on the regression datasets. In general, the prediction results on the test sets suggest that synthetic data produced by CasTGAN is well suited to fitting machine learning models, as the predictive performance is comparable to training on the real data, and in-line with the best performing state-of-the-art tabular GANs. 

\subsection{Univariate Similarity}

\begin{figure*}[!htbp]
    \centering
    \begin{subfigure}{\columnwidth}
        \includegraphics[width=\textwidth]{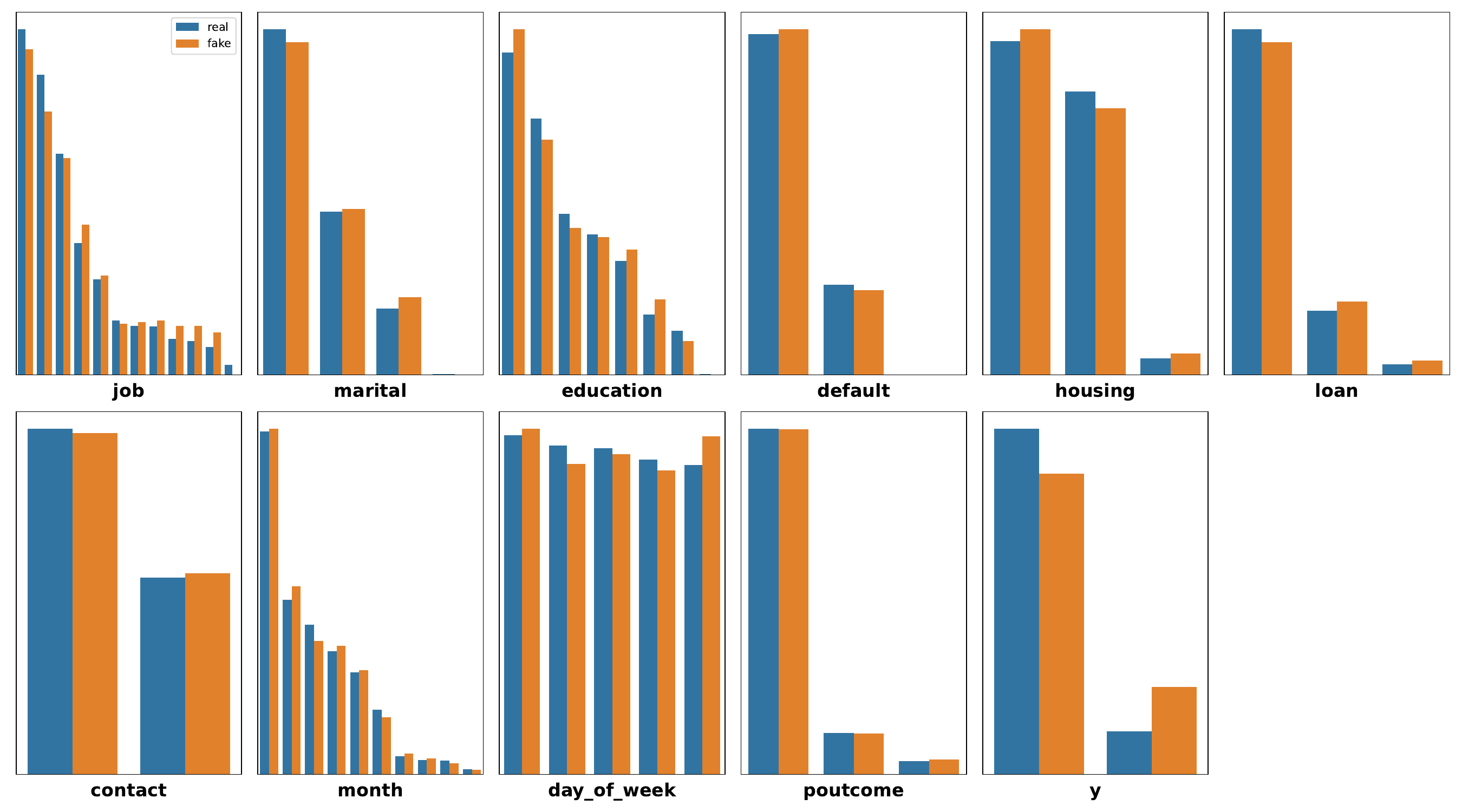}
        \caption{Univariate categorical feature-wise distribution comparison between the real data (in blue) and the CasTGAN synthetic data (in orange). The counts of the categories are log-scaled for intuitively representing infrequent categories.}
        \label{fig:bank_stat_cat}
    \end{subfigure}
    \hfill
    \begin{subfigure}{\columnwidth}
        \includegraphics[width=\textwidth]{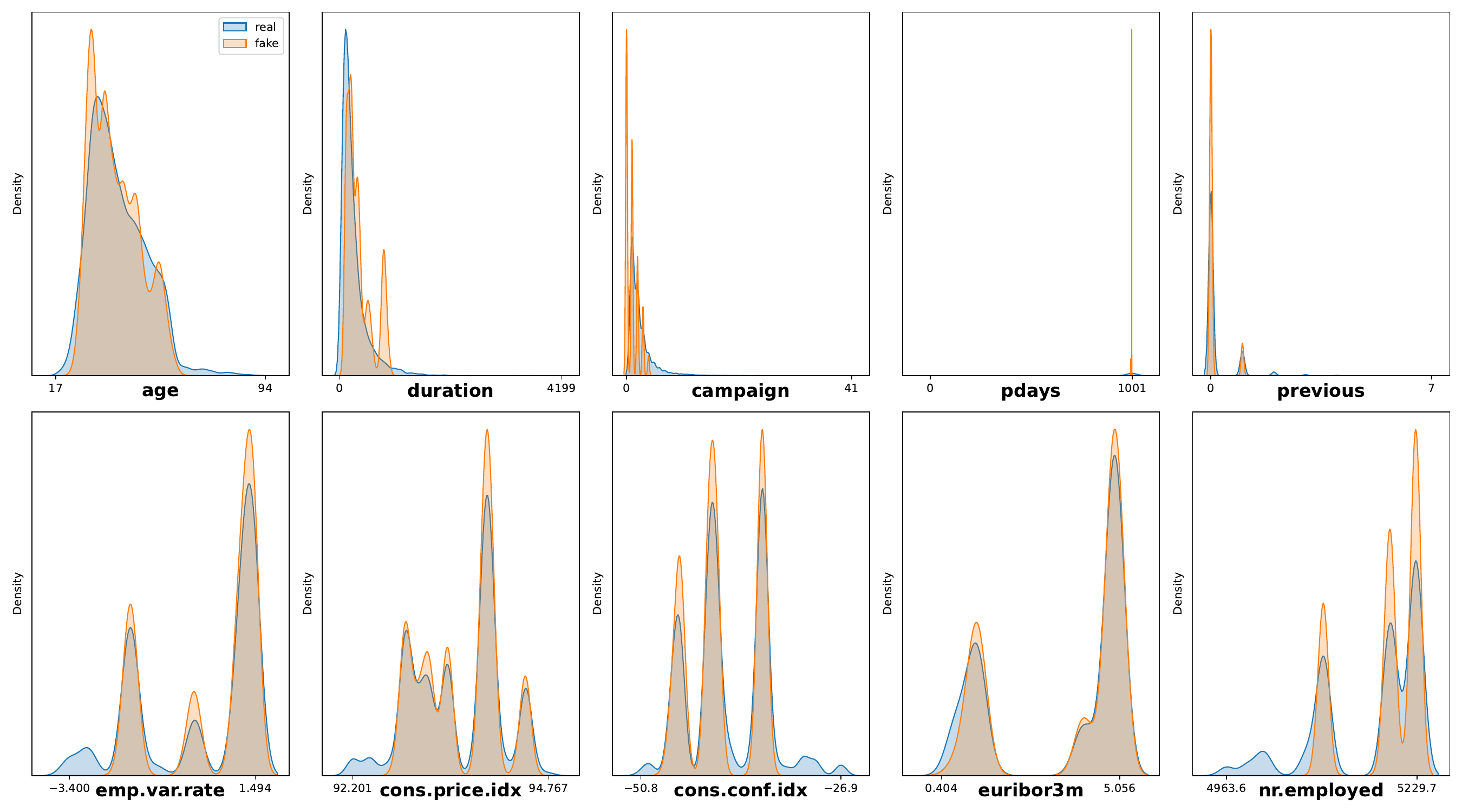}
        \caption{Univariate numerical feature-wise distribution comparison between the real data (in blue) and the CasTGAN synthetic data (in orange). The distributions are plotted by applying Gaussian kernel density estimation.}
        \label{fig:bank_stat_num}
    \end{subfigure}
    \caption{Discrete and continuous univariate features distribution plots for the Bank dataset.}
    \label{fig:stat_sim}
\end{figure*}

\begin{table*}[!htbp]
    \caption{Univariate dimension-wise statistical comparison showing the Euclidean root mean squared error (Euc. RMSE) and the Kolmogorov-Smirnov two-sample test score (KS statistic)}.
     \label{tb:univar}
     \centering
     \vspace*{10pt}
    \begin{subtable}[h]{0.70\textwidth}
        \centering
        \caption{Adult, Bank and Credit datasets}
        \resizebox{\textwidth}{!}{
        \label{tb:univar_a}
        \begin{tabular}{c| cc|cc|cc}
        \toprule
        & \multicolumn{2}{c|}{Adult} & \multicolumn{2}{c|}{Bank} & \multicolumn{2}{c}{Credit} \\
        & Euc. RMSE & KS statistic & Euc. RMSE & KS statistic & Euc. RMSE & KS statistic \\
        \midrule
        table-GAN & 0.2249 & 0.6499 & 0.3485 & 0.5969 &  0.3218 & 0.8794 \\
        medGAN & 0.0599 & 0.1290 & 0.0636 & 0.1963 &  0.0570 & 0.3038 \\
        CTGAN & 0.0377 & 0.1502 & 0.0415 & 0.1076 &  0.0786 & 0.1715 \\
        cWGAN & 0.0406 & 0.1111 &  0.0616 & 0.5536 & \textbf{0.0167} & 0.1906\\
        CTAB-GAN & 0.0186 & \textbf{0.0812} & 0.0396 & \textbf{0.0871} &  0.0335 & \textbf{0.0959} \\
        CasTGAN & \textbf{0.0111} & 0.0908 & \textbf{0.0276} & 0.1470 & 0.0283 & 0.1327 \\
        \bottomrule
       \end{tabular}
        }
    \end{subtable}

    \vspace*{20pt}
    
    \begin{subtable}[h]{0.70\textwidth}
        \centering
        \caption{Diabetes, Cars and Housing datasets}
        \resizebox{\textwidth}{!}{
        \label{tb:univar_b}
        \begin{tabular}{c| cc|cc|cc}
        \toprule
        & \multicolumn{2}{c|}{Diabetes} & \multicolumn{2}{c|}{Cars} & \multicolumn{2}{c}{Housing} \\
        & Euc. RMSE & KS statistic & Euc. RMSE & KS statistic & Euc. RMSE & KS statistic \\
        \midrule
        table-GAN & 0.4087 & 0.4247 & 0.0531 & 0.7088 & 0.1476 & 0.6606 \\
        medGAN & \textbf{0.0188} & \textbf{0.0447} &  0.0124 & 0.1323 & \textbf{0.0141} & 0.1812\\
        CTGAN & 0.0767 & 0.1024 &  0.0067 & 0.0785 & 0.0571 & 0.1682\\
        cWGAN & 0.1133 & 0.2809 & 0.1065 & 0.2890 &  0.1096 & 0.2956 \\
        CTAB-GAN & 0.0648 & 0.0798 &  - & - & 0.0150 & \textbf{0.0989}\\
        CasTGAN & 0.0834 & 0.0780 &  \textbf{0.0019} & \textbf{0.0457} & 0.0184 & 0.1651 \\
        \bottomrule
       \end{tabular}
       }
    \end{subtable}
     
\end{table*}

It is imperative that synthetic data generation techniques need to emulate the distribution of the features of the real data. One method for qualitatively evaluating the statistical similarity between the real data and the synthetic data is to visually compare the distributions for categorical and numerical attributes. Subsequently, we choose to display the comparison between the synthetic and real features for the Bank dataset, as it consists of a diverse and heterogeneous set of features. The depiction is demonstrated in Figure~\ref{fig:stat_sim}.

From Figure~\ref{fig:stat_sim}, we can observe how well CasTGAN performs in approximating the distributions for the categorical and numerical attributes. For categorical variables in Figure~\ref{fig:bank_stat_cat}, it is evident that CasTGAN preserves the frequency of unique categories under each discrete variable. It can also be observed that our framework can successfully represent and sample the less frequent categories in the dataset. For the numerical features in Figure~\ref{fig:bank_stat_num}, we can find that the synthetic data distribution closely approximates the numerical distribution the real data. While adopting the Variational Gaussian Mixture models in CasTGAN helps in improving the density estimation of numerical features, as indicated by the number of peaks and the extensive coverage of the numerical bounds, we notice that approximation is slightly conservative, thus, does not represent extreme values in the real data's numerical range to avoid violating boundary constraints and generating invalid records. The combined results of the numerical and categorical features distributions of the synthetic data from CasTGAN, in addition to the reasonable sampling of less frequent features, as demonstrated in Figure~\ref{fig:stat_sim}, signifies that our framework clearly does not suffer from mode collapse.

To quantitatively analyse how well CasTGAN represents the feature distributions of the original data, we compare the Euclidean distance RMSE and Kolmogorov-Smirnov statistic between the synthetic data and the real data in Table~\ref{tb:univar}. We find that CasTGAN performs considerably the best in terms of the Euclidean distance RMSE on three of the six benchmark datasets: Adult, Bank, and Cars, while ranking comparatively to the best performing GANs on the three remaining datasets. It can also be observed that while CasTGAN only achieves the best KS statistic on the Cars dataset, and often trailing only slightly behind CTAB-GAN on the other datasets, the KS statistic by our model regularly maintains a small value, which can be interpreted as the higher likelihood of CasTGAN's synthetic data to come from the same distribution as the real data. Generally, we note that our CasTGAN, along with medGAN and CTAB-GAN dominate the dimension-wise statistical similarity test. It is also apparent that CasTGAN represents the features of the Adult and Cars datasets particularly well, while performing comparatively on the remaining datasets. We can therefore deduce that CasTGAN is particularly useful for datasets with a greater number of unique categories.

\subsection{Output Validity}

\label{sbs:results_outputvalidity}

\begin{figure*}[!htbp]
    \centering
    \begin{subfigure}{0.92\columnwidth}
        \includegraphics[width=\columnwidth]{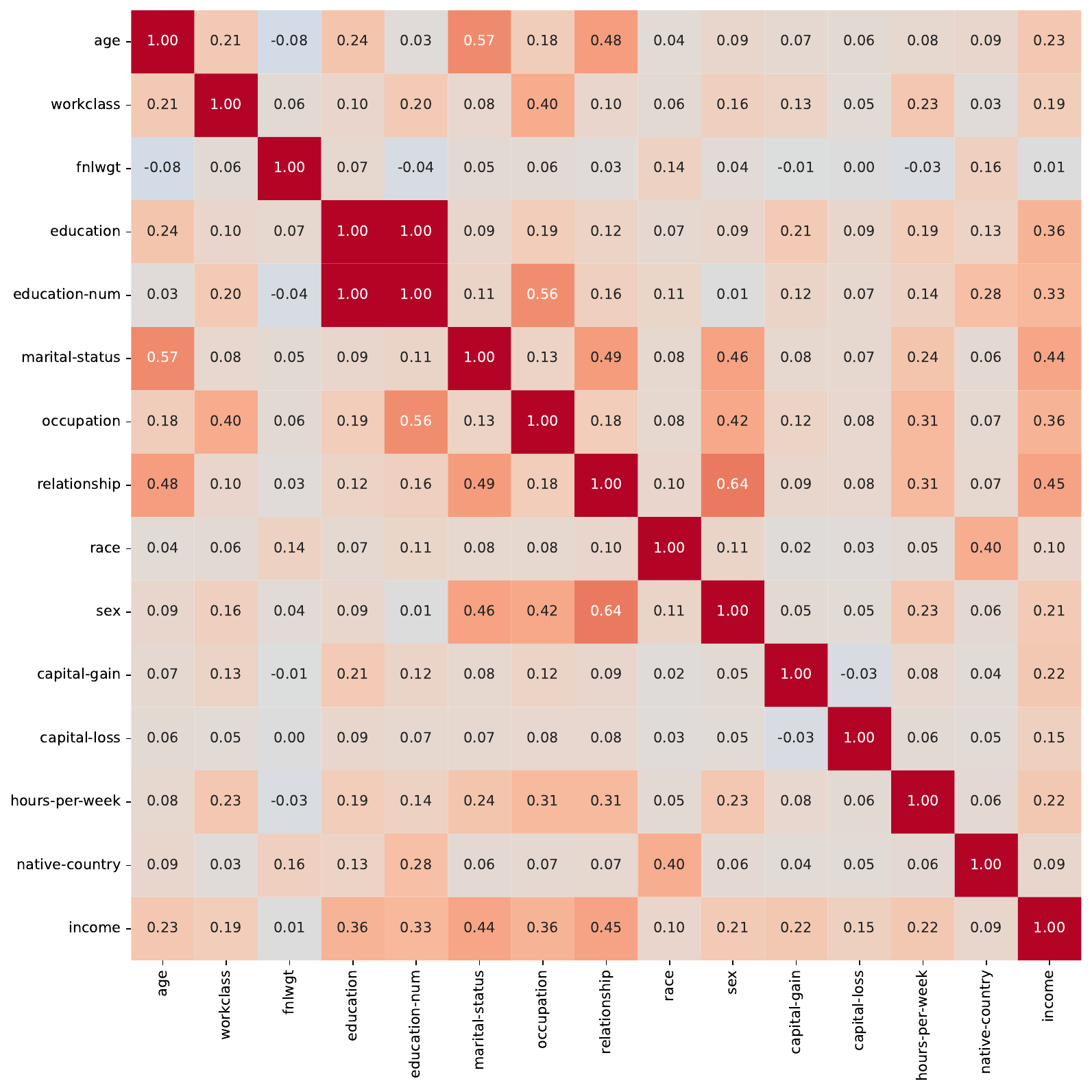}
        \caption{Real data correlation matrix}
        \label{fig:corr_real}
    \end{subfigure}
    \begin{subfigure}{0.92\columnwidth}
        \includegraphics[width=\columnwidth]{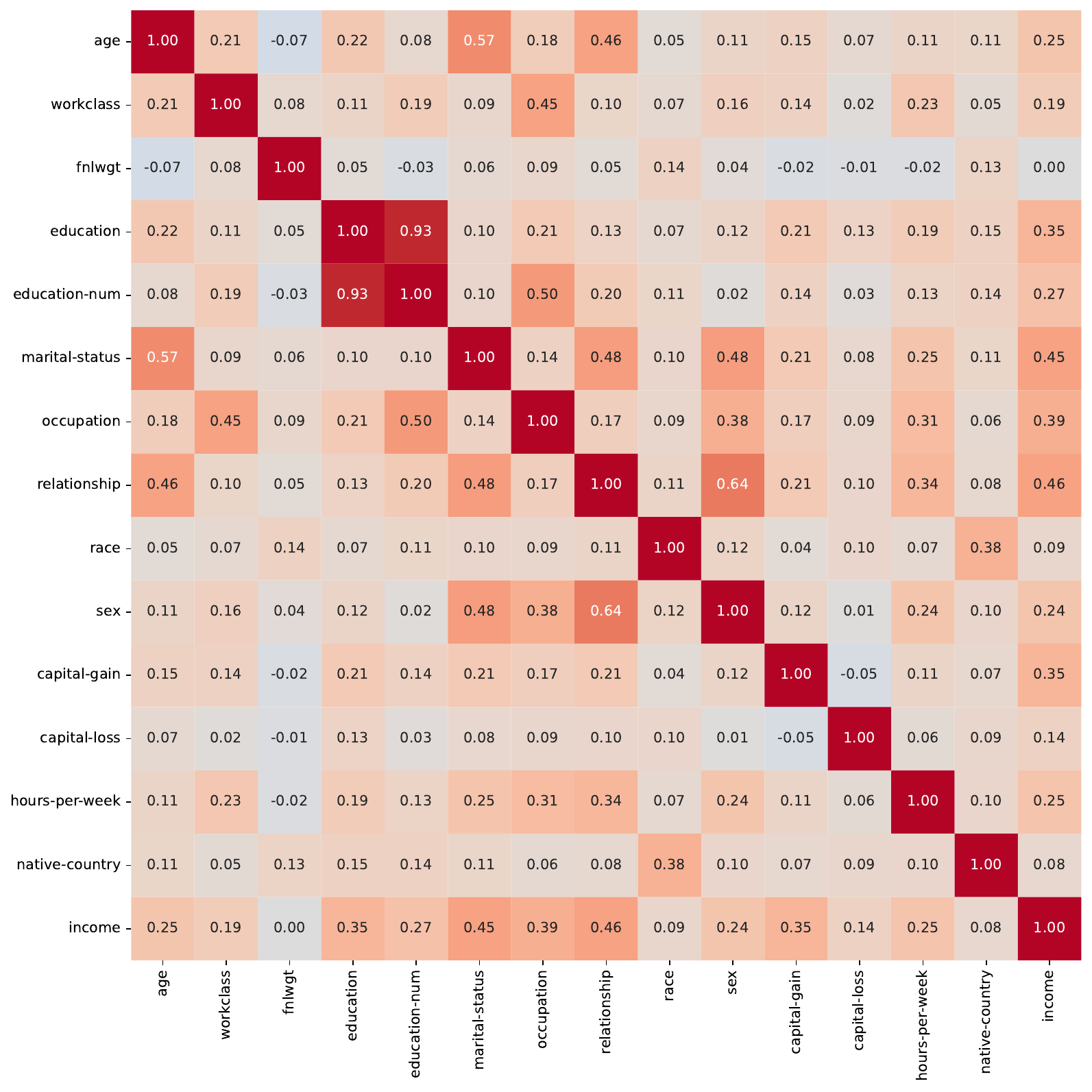}
        \caption{CasTGAN synthetic data correlation matrix}
        \label{fig:corr_syn}
    \end{subfigure}
    \caption{Adult dataset correlation map plots for the real and the synthetic data generated by CasTGAN. Larger absolute values indicate a stronger correlation, either positively or negatively.}
    \label{fig:corr_matrix}
\end{figure*}

In addition to ensuring univariate similarity, it is equally fundamental to evaluate how well the synthetic models preserve the interactions between the different features of a dataset. Inspecting the correlations between the data features of the real data and comparing the correlations with the synthetic data representation can indicate whether the generative data models are capable of simulating the relationship between the data variables from the real data representation. Figure~\ref{fig:corr_matrix} depicts the correlation matrices for the Adult real training data and the synthetic Adult data generated by CasTGAN. We can see from Figure~\ref{fig:corr_matrix} that our model notably learns the correlations between the real data features during the training process. By comparing Figure~\ref{fig:corr_real} and Figure~\ref{fig:corr_syn}, it is evident that there is only a marginal difference between the correlations of the real data and the CasTGAN synthetic data, the highest of which we note is a correlation difference of 0.14 between the "native-country" and "education-num" features. Otherwise, the model successfully emulates both the magnitude of the correlations and the sign, as to whether the feature correlations are positive or negative.

\begin{table*}[htb]
    \caption{Diversity and correlation comparison demonstrating the number of unique pairwise categorical combinations (UPCC), the correlation root mean squared error (Corr. RMSE) and the CORDV score.}
     \label{tb:cordv}
     \centering
     \vspace*{10pt}
    \begin{subtable}[h]{0.70\textwidth}
        \centering
        \caption{Adult, Bank and Credit datasets}
        \resizebox{\textwidth}{!}{
        \label{tb:corr_a}
        \begin{tabular}{c| ccc|ccc|ccc}
        \toprule
        &  \multicolumn{3}{c|}{Adult} & \multicolumn{3}{c|}{Bank} & \multicolumn{3}{c}{Credit}  \\
        & UPCC & Corr. RMSE & CORDV  & UPCC & Corr. RMSE & CORDV  & UPCC & Corr. RMSE & CORDV  \\
        \midrule
        Identity & 3004 &  &  & 1244 &  &  &  1426 &  &  \\
        \hline
        table-GAN & 1168 & 0.2010 & 0.5170 & 461 & 0.2516 & 0.6789 &  878 & 0.3001 &  0.4874  \\
        medGAN & 3769 & 0.1708 & 0.1361 & 1233 & 0.1371 & 0.1383 &  1971 & 0.1826 &  0.1321 \\
        CTGAN & \textbf{3435} & 0.0570 & 0.0498 & \textbf{1306} & \textbf{0.0749} & \textbf{0.0713} &  \textbf{2501} & 0.1003 &  \textbf{0.0572} \\
        cWGAN & 1205 & 0.1899 & 0.4734 & 993 & 0.2043 & 0.2559 &  734 & 0.2612 &  0.5075 \\
        CTAB-GAN & 2404 & 0.0718 & 0.0897 & 1255 & 0.1128 & 0.1118 &  1963 & \textbf{0.0804} &  0.0584  \\
        CasTGAN & 2369 & \textbf{0.0368} & \textbf{0.0466} & 1163 & 0.0822 & 0.0879 &  1191 & 0.1130 &  0.1353 \\
        \bottomrule
       \end{tabular}
        }
    \end{subtable}

    \vspace*{20pt}
    
    \begin{subtable}[h]{0.70\textwidth}
        \centering
        \caption{Diabetes, Cars and Housing datasets}
        \resizebox{\textwidth}{!}{
        \label{tb:corr_b}
        \begin{tabular}{c| ccc|ccc|ccc}
        \toprule
        &  \multicolumn{3}{c|}{Diabetes} & \multicolumn{3}{c|}{Cars} & \multicolumn{3}{c}{Housing}  \\
        & UPCC & Corr. RMSE & CORDV  & UPCC & Corr. RMSE & CORDV  & UPCC & Corr. RMSE & CORDV  \\
        \midrule
        Identity & 420 &  &  & 62872 &  & &  85 &  &  \\
        \hline
        table-GAN & 134 & 0.1482 & 0.4646 & 28 & 0.4015 & 901.5202 & 62 & 0.3493 & 0.7890  \\
        medGAN & \textbf{420} & 0.0891 & 0.0891 & 165522 & 0.2734 & 0.1039 &  70 & 0.2220 &  0.2696 \\
        CTGAN & \textbf{420} & 0.0575 & 0.0575 & \textbf{128405} & 0.1677 & \textbf{0.0821} & \textbf{140} & 0.1755 & 0.1065 \\
        cWGAN & 392 & 0.1135 & 0.1216 & 739 & 0.3409 & 29.0038 & 69 & 0.2322 & 0.2861 \\
        CTAB-GAN & \textbf{420} & 0.0512 & 0.0512 & - & - & - & 107 & 0.1384 & 0.1100 \\
        CasTGAN & 406 & \textbf{0.0490} & \textbf{0.0507} & 31782 & \textbf{0.1351} & 0.2672 & 83 & \textbf{0.0880} & \textbf{0.0909} \\
        \bottomrule
       \end{tabular}
       }
    \end{subtable}
     
\end{table*}

As emphasised in the motivation for this work, there is a need to establish evaluation frameworks that can critically assess the \emph{realistic-ness} of the synthetically generated data. We therefore quantify the validity of the output by considering the number of unique pairwise categorical combinations (UPCC), the correlation error between the synthetic and the real data, and the CORDV score, which is essentially the correlation divided by the UPCC ratio. From Table~\ref{tb:cordv}, we can observe how the different generative methods rank among the three aforementioned metrics. First, we note that CTGAN and medGAN perform well in generating a large number of unique feature combination, in most cases, even more than the number of combinations that can be found in the training data. This is particularly impressive for the Cars dataset, where both models managed to generate more than twice the number of categorical combinations of the training set. Similarly, we also observe that CTAB-GAN performs relatively well in exploring diverse categorical combinations. Meanwhile, we notice that our CasTGAN is more conservative when it comes to generate unique categorical combinations. For all the datasets, CasTGAN produces a marginally lower number of pairwise combinations than can be typically found in the training set. Moreover, the UPCC can be a good indicator of mode collapse and this is reflected by the significantly low UPCC values for table-gan and cWGAN, where it can be deduced that these models generated a limited number of modes for some categories.

In contrast, it can be observed from Table~\ref{tb:cordv} that CasTGAN generally outperforms the other baselines in capturing the feature correlations of the datasets. The lower correlation RMSE score entails that CasTGAN prioritises the representation of correlations and feature interdependence in the real data. Meanwhile, the CORDV score aims to quantify the trade-off between the diversity and the proximity of the synthetic data to the real data. We observe from the results that the highest CORDV scores are evenly split among our CasTGAN and CTGAN.

Despite lacking the full domain knowledge in our datasets, we nonetheless measure the synthetic invalidity on the Adult, Cars and Housing datasets as follows:
\begin{itemize}
    \item \emph{Adult dataset}: we use the fields "relationship" and "sex" to calculate the number of invalid records. We posit that if "relationship" = "Husband", then the sex feature needs to be set to "Male". Likewise, the "relationship" = "Wife" needs to align the gender field assigned as "Female". This is not based on our assumptions, but rather running exploratory data analysis on the training set confirms that the records are matched in such a manner.
    \item \emph{Cars dataset}: we use the fields "make" and "model" and classify a synthetic sample as invalid if the synthetic car's "model" does not in fact belong to the "make" that can be found in training data.
    \item \emph{Housing dataset}: the housing dataset consists of the fields "year built" and "year renovation". Logically, a property cannot be renovated before it was built, and further inspecting the data indeed confirms that there are no observations with "year renovation" that precedes "year built". 
\end{itemize}
Based on the aforementioned fronts, the ratio of invalid records generated by CasTGAN and three baseline methods are demonstrated in Table~\ref{tb:invalidity}. We chose not to include table-GAN and cWGAN in the comparison as their synthetic output was found to be characterised by major mode collapse.

\begin{table}[htb]
\caption{An outline of the ratio of invalid synthetic records of the baseline GANs and our model.}
\centering
\label{tb:invalidity}
    \begin{tabular}{l ccc}
        \toprule
        &  Adult &  Cars & Housing\\
        \midrule
        medGAN & 29.05\% & 91.11\% & 59.29\%\\
        CTGAN & 9.82\% & 42.89\% & 54.75\%\\
        CTAB-GAN & 4.14\% & - & 44.42\%\\
        CasTGAN & \textbf{0.67\%} &  \textbf{27.82\%} & \textbf{22.98\%}\\
        \bottomrule
    \end{tabular}
\end{table}

From Table~\ref{tb:invalidity} we can observe that CasTGAN remarkably reduced the number of invalid synthetic records of the Adult dataset. Our method also significantly decreased the number of invalid records in the Cars dataset. This resembles a major improvement from CTGAN, while noting the challenging nature of modelling the Cars dataset due to the large number of categories present. Furthermore, it is evident that CasTGAN outperforms the other generative approaches on the numerical features of the Housing dataset. 

\subsection{Robustness Against Privacy Attacks}

For conducting white-box privacy attacks, we set the number of attacking iterations to five, where each feature in the synthetic data is updated five times based on the output of the auxiliary learners. Moreover, the membership attacks are launched on 10\% of the total number of overall samples. We highlight that the ratio of attacked samples does not impact the evaluation of the robustness of our approach as we only compute the attack distance metrics with respect to the attacked samples. The Euclidean distance of the attacked samples to the training data and to the synthetic data prior to the membership attacking is computed in Table~\ref{tb:wbb}.

\begin{table*}[!htbp]
\small
\centering
\caption{White box privacy attacks on the auxiliary learners proximity measures.}
\label{tb:wbb}
\resizebox{0.70\linewidth}{!}{
\begin{tabular}{cc| cc | cc}
    \toprule
    & & \multicolumn{2}{c}{Access to only AL} & \multicolumn{2}{|c}{Access to AL and Preprocessing} \\
    & & Euc. to train & Euc. to pre-attack syn &  Euc. to train & Euc. to  pre-attack syn \\
    \midrule
    
    \multirow{4}{*}{Adult} & $\epsilon = 0.0$ & 1.2202 & 1.4437 & 0.1256 & 0.3652\\
    & $\epsilon = 0.1$ & 1.4250 & 1.7289 & 0.2025 & 0.7348\\
    & $\epsilon = 0.2$ & 1.4747 & 1.8288 & 0.6848 & 1.4039\\
    & $\epsilon = 0.3$ & \textbf{2.4468} & 2.5217 & \textbf{0.7413} & 1.2509\\
    \hline
    
    \multirow{4}{*}{Bank} & $\epsilon = 0.0$ & 2.2275 & 2.4975 & 1.2225 & 1.7366\\
    & $\epsilon = 0.1$ & 3.0541 & 3.0272 & 2.4020 & 2.4182\\
    & $\epsilon = 0.2$ & 5.6174 & 5.9515 & 5.3565 & 5.8131\\
    & $\epsilon = 0.3$ & \textbf{6.7180} & 7.1127 & \textbf{6.7095} & 7.0654\\
    \hline
    
    \multirow{4}{*}{Credit} & $\epsilon = 0.0$ & 0.3740 & 0.6550 & 0.1157 & 0.3179\\
    & $\epsilon = 0.1$ & 0.6196 & 0.7916 & 0.1094 & 0.2804 \\
    & $\epsilon = 0.2$ & 0.3918 & 0.5797 & 0.0732 & 0.2773 \\
    & $\epsilon = 0.3$ & \textbf{1.5533} & 1.7834 & \textbf{0.6349} & 0.8470 \\
    \hline
    
    \multirow{4}{*}{Housing} & $\epsilon = 0.0$ & 0.3696 & 0.4905 & 0.2026 & 0.4132 \\
    & $\epsilon = 0.1$ & 4.5976 & 4.7431 & 4.4731 & 4.5013 \\
    & $\epsilon = 0.2$ & 16.2106 & 16.6204 & 16.1899 & 16.5996 \\
    & $\epsilon = 0.3$ & \textbf{28.6935} & 28.8148 & \textbf{28.6990} & 28.8205 \\
    \bottomrule
\end{tabular}
}

\end{table*}

From Table~\ref{tb:wbb}, it can be observed that the perturbation coefficient $\epsilon$ greatly impacts the closeness of attacked samples to the training data. For unperturbed and minimally perturbed data features, it can be noticed that the attacked samples are relatively close to the training samples, indicating that the attackers might succeed in recovering training datapoints. We observe that the proximity to the training samples increases for greater $\epsilon$ values, which demonstrates the additional privacy guarantees that can be provided when altering the labels. We additionally notice how access to the data processors gives a major advantage to the attackers attempting to recover the training samples. This holds true especially for the Adult dataset, where using the trained label encodings of the auxiliary learners lead to more targeted attacks that are even closer to the training samples than attackers on unperturbed data with no access to the data preprocessors. Another interesting observation is that the attacks on the Housing dataset are greatly impacted by incremented in $\epsilon$, which is plausible, given that the dataset mainly consists of numerical features.

In addition to the proximity to the training samples, we also investigate whether perturbing the labels of the auxiliary learners can contribute to a reduction in the quality of unattacked synthetic data as demonstrated in Table~\ref{tb:wbb_utilt}. For the Adult, Bank and Credit datasets it can be evident that applying perturbations insignificantly impacts the the evaluation metrics of the synthetic datasets. We observe that the PR-AUC scores on the test data and KS statistic for univariate distributions are minimally influenced by the changes in $\epsilon$. In contrast, it appears that perturbing the data impacts the CORDV scores as a result of the correlation errors between the synthetic and the real datasets. Interestingly, applying perturbations on the Housing dataset appear to improve the quality of the synthetic output in addition to increasing its robustness against white-box privacy attacks.

\begin{table}[!tb]
\small
\centering
\caption{Impact of auxiliary learners imputation on the synthetic output.}
\label{tb:wbb_utilt}
\begin{tabular}{cc| ccc}
    \toprule
    & & TSTR & KS & CORDV \\
    \midrule
    
    \multirow{4}{*}{Adult} & $\epsilon = 0.0$ & 0.6718 & \textbf{0.0908} & 0.0467\\
    & $\epsilon = 0.1$ & 0.6683 & 0.1173 & \textbf{0.0380}\\
    & $\epsilon = 0.2$ & \textbf{0.6731} & 0.1560 & 0.0393\\
    & $\epsilon = 0.3$ & 0.6647 & 0.1346 & 0.0425\\
    \hline
    
    \multirow{4}{*}{Bank} & $\epsilon = 0.0$ & \textbf{0.5657} & 0.1470 & 0.0879\\
    & $\epsilon = 0.1$ & 0.5384 & \textbf{0.1020} & \textbf{0.0847}\\
    & $\epsilon = 0.2$ & 0.5551 & 0.1253 & 0.0868\\
    & $\epsilon = 0.3$ & 0.5475 & 0.1130 & 0.0960\\
    \hline
    
    \multirow{4}{*}{Credit} & $\epsilon = 0.0$ & 0.4995 & 0.1327 & \textbf{0.1353}\\
    & $\epsilon = 0.1$ & \textbf{0.5140} & \textbf{0.1069} & 0.1504\\
    & $\epsilon = 0.2$ & 0.4722 & 0.1170 & 0.1434\\
    & $\epsilon = 0.3$ & 0.5095 & 0.1103 & 0.1822\\
    \hline
    
    \multirow{4}{*}{Housing} & $\epsilon = 0.0$ & -0.4430 & 0.1651 & 0.0909\\
    & $\epsilon = 0.1$ & -0.5602 & 0.2185 & 0.0864\\
    & $\epsilon = 0.2$ & -0.3257 & 0.1554 & 0.0961\\
    & $\epsilon = 0.3$ & \textbf{-0.2966} & \textbf{0.1381} & \textbf{0.0828}\\
    \bottomrule
\end{tabular}

\end{table}


\subsection{Impact of Auxiliary Learners Loss}

\begin{figure}[!htbp]
    \centering
    \includegraphics[width=\columnwidth]{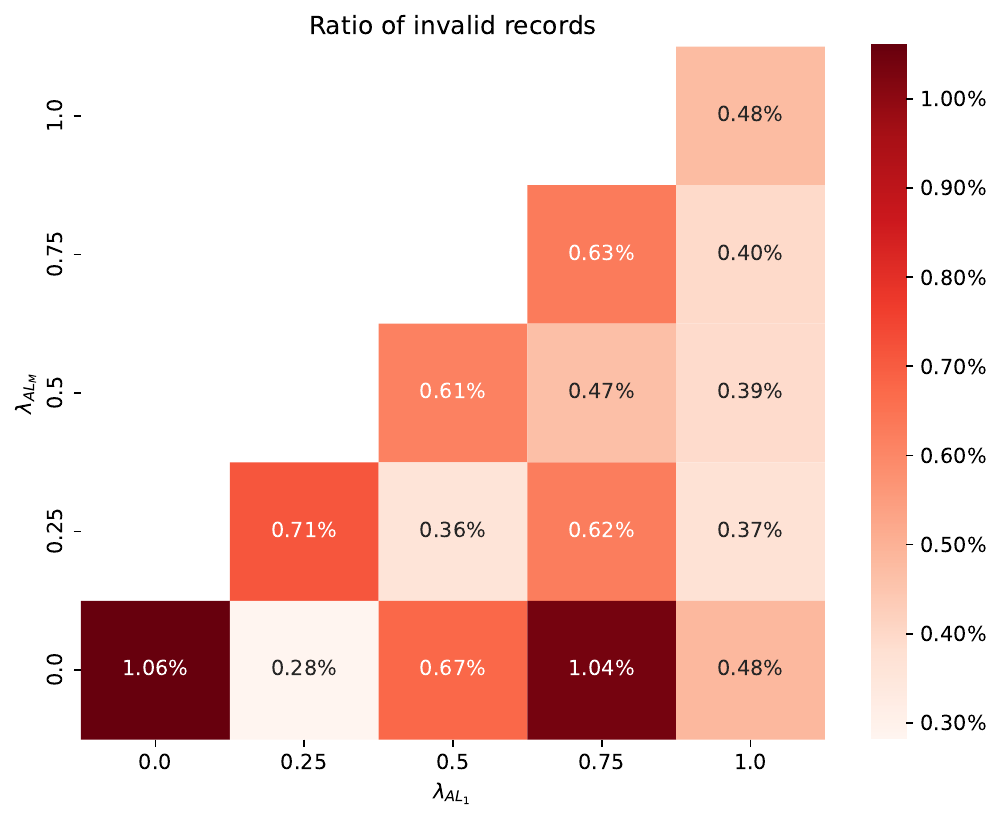}
    \caption{Proportion of invalid synthetic records of the Adult dataset generated by CasTGAN using various auxiliary loss coefficient settings.}
    \label{fig:aux_hyper}
\end{figure}

We further analyse the impact of the auxiliary learners on the quality of the synthetic data samples produced by CasTGAN. To this end, we conduct experiments on our model by tuning the auxiliary loss coefficient parameters $\lambda_{AL_1}, \lambda_{AL_2}, \ldots, \lambda_{AL_M}$. This is implemented by adjusting $\lambda_{AL_1}$ and $\lambda_{AL_M}$, as the auxiliary loss coefficients in between are linearly and equidistantly scaled between the loss coefficients of the first and final auxiliary learners. Given that there is an indefinite number of auxiliary loss coefficients combinations and a diverse set of evaluation metrics and datasets, we conduct our the analysis on the Adult dataset by experimenting with a set of auxiliary loss coefficient values: $\{0, 0.25, 0.5, 0.75, 1\}$. As the focus of our work is improving the \emph{realistic-ness} and maximising the number of semantically valid synthetic records, we quantify the number of invalid records of the synthetic samples as per the description of invalid records of the Adult dataset defined in Section \ref{sbs:results_outputvalidity}. Figure~\ref{fig:aux_hyper} demonstrates the fraction of invalid records generated by CasTGAN for various auxiliary loss coefficients.

As shown in Figure~\ref{fig:aux_hyper}, adjusting the auxiliary loss coefficients is reflected by a modest change in the ratio of invalid synthetic records of the Adult dataset. As demonstrated, it appears that using a fixed auxiliary loss coefficient across the auxiliary learners by setting the value $\lambda_{AL_1}$ and $\lambda_{AL_M}$ generally increases the number of invalid records. Meanwhile, it can be observed that having a smaller difference between $\lambda_{AL_1}$ and $\lambda_{AL_M}$ leads to an improved performance in contrast to when the difference is $0.5$ or greater in general. We can notice that the lowest fraction of invalid synthetic observations is obtained by setting $\lambda_{AL_1}=0.25$ and $\lambda_{AL_M}=0$. By comparing the ratio of invalid records in Table \ref{tb:invalidity} and Figure \ref{fig:aux_hyper}, it is clear that the default auxiliary loss coefficient parameters that we used in our experimental setup do not yield the most optimal results, whereas further hyperparameter tuning can in practice help reduce the number of invalid records. Meanwhile, it is evident that not relying on auxiliary learners by setting the loss coefficients to zero contributes to the highest fraction of invalid synthetic records. However, we can observe that the ratio of invalid records generated when using loss coefficients of $0$, thereby nullifying the effect of auxiliary learners, is nevertheless lower than the proportion of the invalid synthetic records generated by the baseline models, as displayed in Table \ref{tb:invalidity}. This suggests that while the improved quality of synthetic records can be to a great extent be attributed to the design of the cascaded generator architecture and employment of the WGAN-GP, the use of auxiliary learners and fine-tuning the auxiliary loss coefficients can further contribute to the generation more semantically valid records.


\section{Discussion}

\label{sc:discussion}

Data in tabular form is widely used across organisations in various domains for decision support systems. In such systems, data mining techniques and statistical models are employed to enhance the predictability of significant events and to gain a deeper insight into user behaviour. Moreover, mixed-type tabular data is commonly used for facilitating knowledge exchange between domain experts and stakeholders. Given the privacy concerns associated with sharing confidential information, in addition to the scarcity of available data to share across organisations, there has been a growing interest in tabular data synthesis using generative adversarial networks, in which realistic synthetic data can be safely shared across operational units. In literature, the focus of generative models has been primarily directed towards generating synthetic data that yields acceptable machine learning utility and exhibits statistical properties which are similar to the real data. While these objectives are relatively significant, an underlying limitation is that the semantic integrity of the synthetic output is not thoroughly examined. Given that GANs attempt to approximate the data distributions loss minimising functions for the generator and the discriminator, it is imperative that GAN-based synthesis models are unable explicitly distinguish between semantically valid and invalid data records.

To this end, the cascaded tabular GAN architecture we propose in this study contributes to the reduction of the number of invalid generated observations by dedicating a generator for every feature of the dataset. The sequential cascading of generator passes incomplete synthesisation of the data to subsequent generators, in which each generator attempts to predict its target feature from the incomplete feature space it receives from the preceding generator. As the generator fills the remaining feature space with noise, the output is queried against the discriminator and the discriminator loss for the queried sample is backpropagated to the generator for improving its overall generation. For obtaining meaningful feedback to the generator's designated feature, an auxiliary learner is coupled to the generator that calculates the prediction loss for the specific feature and propagates the loss to the generator, such that it can learn to reduce the auxiliary learner's loss during GAN training. In such an architecture, the GAN attempts to learn the dependencies between the dataset's features, and thereby gradually reduces the losses for the components of the GAN during the training process. Hence, the aim of CasTGAN is to increase the semantic integrity of the synthetic and reduce the invalid records that are generated. This is highly significant as the generation of invalid records may have adverse effects on knowledge sharing, potentially fostering an inaccurate comprehension of the data among stakeholders and other external entities granted access to the synthesized data.

Notably, our results indicate where our proposed generative model has remarkably performed. First, the machine learning utility results demonstrate that the synthetic output from CasTGAN is particularly useful for constructing predictive models, as reflected by the classification and regression performance metrics. The results show that CasTGAN is competitive with the state-of-the-art tabular GANs, outperforming them on some datasets, and barely falls short to the predictive performance obtained by training on the real data. Regarding the univariate variable properties, we visually demonstrate the synthetic output has a striking similarity to the the real data numerical and categorical variables. Furthermore, it can be observed that quantitative similarity analysis indicates a very close similarity between the data distributions of the real and the CasTGAN synthetic output. In accordance to addressing the gap in literature for better exploring the semantic integrity of the synthetic data, we evaluate the output validity of our synthetic data on several fronts. The correlation mapping shows that the synthetic data from our model exhibits a strong resemblance to the real data used for training. We also notice that CORDV metric which we design for measuring the correlation error as a fraction of the unique pairwise combination successfully shows that our approach is well capable of capturing the dependencies between the data features. In general, the performance of CasTGAN for machine learning usability, statistical similarity and correlation and diversity was mostly comparable with CTGAN and CTAB-GAN. This can be attributed to the use of WGAN-GP in the case of CTGAN, or classification loss and information loss in the case of CTAB-GAN, both of which contribute to the training stability and the mitigation against mode collapse.  Meanwhile, calculating the ratio of invalid records on a number of datasets demonstrates that CasTGAN excels in reducing the number of invalid data observations in comparison to the existing tabular GAN approaches, thereby improving the potential of using the synthetic data for knowledge exchange. Further analysis unveils that applying perturbations on the auxiliary learners can increase the robustness of our model against privacy attacks without notably sacrificing our model's synthetic output quality.

We also note that using the auxiliary learners leads to a more conservative approach for the GAN training process. The reduction of the invalid records comes at the expense of reducing the number of unique pairwise combinations of the data categories that CasTGAN can synthesise, and additionally not fully simulating numerical values around the boundaries for numerical features. Nevertheless, we highlight that the reduction of invalid records and improving the model's representation of feature dependency was the focus of this work, and that CasTGAN demonstrates success in these two aspects.


\section{Conclusion}

\label{sc:conclusion}

In this work, we presented CasTGAN as a generative framework for creating synthetic tabular data samples that are representative of the real data attributes. Our motivation for this work stems from the need for realistic tabular data that can be exchanged amongst experts, while focusing on the reliability and the sensitivity of such information. We therefore directed our focus towards generating fake output that capture the correlations and interdependence between the data features. We demonstrated that our cascaded generator architecture supported by auxiliary learners are able to generate realistic output given highly dimensional and largely imbalanced tabular datasets. Our results indicate that CasTGAN is capable of significantly reducing the number of invalid records while exhibiting strong statistical and correlational similarities to the real data. We further evaluated the robustness of our model against targeted privacy attacks and showed that perturbing the auxiliary learners by a small scale can mitigate against attacks aiming to recover the real data samples.

Given the challenging nature of generating realistic synthetic tabular data, there are several paths for future work. This work can be extended by incorporating additional data types within the tabular data generation such as free text and timestamps. Due to cascaded architecture and the presence of multiple auxiliary learners, we point out that our framework does not offer improvements to the training speed over the existing tabular GAN models, which presents a potential opportunity for future optimisation efforts. Moreover, we intend to explore how our approach can generate more diversified combination of categories, while maintaining its ability in minimising the number of invalid data records.

\appendices
\section{Implementation Details}

\subsection{\break Hyperparameters}
\label{app:def_hyper}

\begin{table}[!htbp]
\caption{CasTGAN hyperparameters used for all datasets.}
\label{tb:app_hyper}
\small
\centering
\resizebox{0.95\columnwidth}{!}{
    \begin{tabular}{l l}
        \toprule
        Hyperparameter & Value \\
        \midrule
        Epochs & 300 \\
        Batch size & 512 \\
        Generators noise input dimension & 128 \\
        Generators noise input distribution & $\sim \mathcal{N}(0, 1)$ \\
        Generators primary networks hidden sizes & [128, 64] \\
        Generators primary networks inter-hidden layer activations & LeakyReLU \\
        Generators primary networks activation normalizer & Layer Normalization \\
        Generators primary networks numerical feature activation layer & tanh \\
        Generators primary networks categorical feature activation layer &  gumbel softmax$(\tau = 0.8)$ \\
        Generators secondary networks hidden sizes & [32] \\
        Generator optimizer & Adam($\alpha$ = 2 $\times$ 10\textsuperscript{-4}, $\beta_1$ = 0.50, $\beta_2$ = 0.99) \\
        Discriminator primary networks hidden sizes & [256, 128] \\
        Discriminator inter-hidden layer activations & ReLU \\
        Discriminator activation normalizer & Layer Normalization \\
        Discriminator optimizer & Adam($\alpha$ = 2 $\times$ 10\textsuperscript{-4}, $\beta_1$ = 0.50, $\beta_2$ = 0.99) \\
        Discriminator real input noise perturbation & $\sim \mathcal{N}(0, 0.01)$ \\
        Discriminator updates per generators update & 1 \\
        Auxiliary learners boosting iterations & 150 \\
        Auxiliary learners number of leaves & 31 \\
        Auxiliary learners learning rate & 0.10 \\
        Auxiliary learners early stopping & 10 rounds \\
        Wasserstein loss gradient penalty $\lambda_{GP}$ & 10 \\
        Auxiliary learner 1 loss coefficient $\lambda_{AL_1}$ & 0.75 \\
        Auxiliary learner $M$ loss coefficient $\lambda_{AL_M}$ & 0.10 \\
        \bottomrule
    \end{tabular}
}

\end{table}

\subsection{Strategies for Hyperparameter Configuration}

\label{app:hyper_guide}

There are various hyperparameters in CasTGAN, and therefore selecting the optimal hyperparameters is a non-trivial task. Defining what is considered as optimal also poses a challenge, as there may be a trade-off between enhancing the suitability of synthetic output for machine learning model fitting and increasing the semantic validity of the synthetic data. In light of this, we outline the parameters that have some significant impact on training our GAN model, and we describe how they can be tuned for optimising the performance. The parameters with less significance in correspondence to the quality of the synthetic data can utilise the default values from Appendix \ref{app:def_hyper}.

\noindent\textbf{Epochs:} Tuning the number of training iterations can offer improvements to the quality of the synthetic output. For smaller datasets, it can be sufficient to train for a smaller number of iterations, in which the training completes faster. Meanwhile, larger datasets may require a greater number of epochs to demonstrate an improved synthetic output. It is recommended to train for a minimum of 100 epochs, though training the GAN for too long can lead to the memorisation of records by the GAN, inducing potential data records leakage and susceptibility to privacy attacks. 

\noindent\textbf{Batch size:} Batch size refers to the number of samples stored in memory for updating the internal model parameters at every training iteration. A larger batch size is more suitable for datasets with a greater number of features and categories, whereas datasets with fewer features can be sufficiently trained with a smaller batch size. It is worth noting that a larger batch size contributes to a relatively slower training of the GAN and can contribute to a poor model convergence if not trained for a sufficient number of epochs.

\noindent\textbf{Generators noise input dimension:} The optimal size of the noise vector used as the input for the generators is best determined through experimentation. Wasserstein GANs commonly exhibit a higher resolution of the synthetic output through the utilisation of larger noise vector \cite{padala2021effect}, though the significance of input noise size is largely dependent on the data and the other model parameters.

\noindent\textbf{Generators primary networks hidden sizes:} The number and the sizes of the hidden layers can significantly impact the quality of the synthetic output from the generators. Despite designating a generator for each data feature, the generators employ a common hidden layer configuration rather than a specific configuration for each generator. As each generator is responsible for generating one primary feature, the linear layers do not need to be overly sophisticated.

\noindent\textbf{Discriminator primary networks hidden sizes:} The architecture of the discriminator should be able to foster the competitiveness between the discriminator and the generator, hence maintain the adversary throughout the GAN training. If the neural network architecture is relatively simpler than that of the generators, the discriminator will poorly distinguish between the synthetic and the real samples. Meanwhile, a complex discriminator architecture might suppress the learning of the generators, thus, degrade the synthetic output quality of the generators. Therefore, the number and the size of the discriminator's hidden layers should be comparable to those employed by the generators.

\noindent\textbf{Discriminator updates per generators update:} It might be worthwhile to adjust the number of discriminator iterations for each generator iteration, if it helps in converging the losses during training. This might be particularly the case if the discriminator is visibly weaker than the generator. If there is no major disparity between the discriminator's architecture and the generators' hidden layers, then there is no proven benefit in employing more than 1 discriminator gradients update for each step of the generators training \cite{goodfellow2016nips}.

\noindent\textbf{Wasserstein loss gradient penalty ($\lambda_{GP}$)}: We find that using a Wasserstein loss gradient penalty of 10 works well, based on the training stability during our preliminary model conceptualisation. This is supported by the authors of the original WGAN-GP paper \cite{gulrajani2017improved}, who found the gradient penalty of 10 works well on various datasets and GAN architectures. This is also in accordance to the baseline tabular GANs with WGAN-GP we compare our approach against \cite{engelmannConditionalWassersteinGANbased2021} \cite{xuModelingTabularData2019}, whom also employ $\lambda_{GP} = 10$.

\noindent\textbf{Auxiliary learner 1 loss coefficient ($\lambda_{AL_1}$)}: The first auxiliary learner in the cascaded generator is assigned with the largest loss coefficient among  to ensure that the univariate statistical representation of the first features of the data conforms to the real data distributions. The particular value for the first auxiliary learner's coefficient is chosen such that large prediction errors attributed to the deviation in distributions are sufficiently penalised, without making it too large such that training instability is introduced to the generators.

\noindent\textbf{Auxiliary learner M loss coefficient ($\lambda_{AL_M}$):} The final auxiliary learner can benefit from low loss coefficient values to minimise the errors propagated to the preceding generators. The purpose of the auxiliary loss coefficient of auxiliary learner M is to reduce the correlation error in the synthetic data towards the end of cascaded generator, as opposed to the first generators that prioritise the univariate similarity. In a similar manner to the first auxiliary learner, this is a hyperparameter that is largely dependent on the dataset used for training and sampling of CasTGAN. It is however advisable to set $\lambda_{AL_M}$ with a very small value if a small value is selected for $\lambda_{AL_1}$.


\subsection{Train on Synthetic, Test on Real}

\begin{table}[!htbp]
  \caption{Adult Dataset Classification}
  \label{tab:app_tstr_adult}
  \centering
  \resizebox{0.85\columnwidth}{!}{
  \begin{tabular}{l|rrrr}
    \toprule
    {} & {Accuracy} & {ROCAUC} & {F1-score} & {PR-AUC} \\
    \midrule
    Identity & 0.8550 & 0.9075 & 0.6663 & 0.7744 \\
    table-GAN & 0.5617 & 0.4102 & 0.1376 & 0.2225 \\
    medGAN & 0.7920 & 0.8038 & 0.4888 & 0.5777 \\
    CTGAN & 0.8304 & 0.8751 & 0.6235 & 0.6932 \\
    cWGAN & 0.7586 & 0.5887 & 0.0002 & 0.3030 \\
    CTAB-GAN & 0.8400 & 0.8849 & 0.6542 & 0.7192 \\
    CasTGAN & 0.8272 & 0.8779 & 0.6380 & 0.6718 \\
    \bottomrule
    \end{tabular}
  }
 \end{table}%

 \begin{table}[!htb]
  \caption{Bank Dataset Classification}
  \label{tab:app_tstr_bank}
  \centering
  \resizebox{0.85\columnwidth}{!}{
  \begin{tabular}{l|rrrr}
  \toprule
    {} & {Accuracy} & {ROCAUC} & {F1-score} & {PR-AUC} \\
    \midrule
    Identity & 0.9084 & 0.9342 & 0.4978 & 0.6085 \\
    table-GAN & 0.2149 & 0.3910 & 0.1670 & 0.0954 \\
    medGAN & 0.8803 & 0.6492 & 0.1547 & 0.2729 \\
    CTGAN & 0.8979 & 0.8474 & 0.3492 & 0.4805 \\
    cWGAN & 0.8337 & 0.6864 & 0.1297 & 0.2692 \\
    CTAB-GAN & 0.8970 & 0.8768 & 0.3550 & 0.4920 \\
    CasTGAN & 0.8991 & 0.9187 & 0.5823 & 0.5657 \\
    \bottomrule
    \end{tabular}
  }
 \end{table}%

\begin{table}[!htb]
  \caption{Credit Dataset Classification}
  \label{tab:app_tstr_credit}
  \centering
  \resizebox{0.85\columnwidth}{!}{
  \begin{tabular}{l|rrrr}
  \toprule
    {} & {Accuracy} & {ROCAUC} & {F1-score} & {PR-AUC} \\
    \midrule
    Identity & 0.8201 & 0.7685 & 0.4663 & 0.5374 \\
    table-GAN & 0.2185 & 0.4605 & 0.3482 & 0.2084 \\
    medGAN & 0.6807 & 0.6146 & 0.3448 & 0.3050 \\
    CTGAN & 0.7826 & 0.6845 & 0.4561 & 0.4558 \\
    cWGAN & 0.7760 & 0.5253 & 0.0053 & 0.2166 \\
    CTAB-GAN & 0.8138 & 0.7136 & 0.4708 & 0.4911 \\
    CasTGAN & 0.8048 & 0.7315 & 0.2845 & 0.4995 \\
    \bottomrule
    \end{tabular}
  }
\end{table}

 \begin{table}[!htb]
  \caption{Diabetes Dataset Classification}
  \label{tab:app_tstr_diabetes}
  \centering
  \resizebox{0.85\columnwidth}{!}{
  \begin{tabular}{l|rrrr}
  \toprule
    {} & {Accuracy} & {ROCAUC} & {F1-score} & {PR-AUC} \\
    \midrule
    Identity & 0.8623 & 0.8142 & 0.2545 & 0.3980 \\
    table-GAN & 0.8604 & 0.5222 & 0.0000 & 0.1585 \\
    medGAN & 0.8519 & 0.7544 & 0.2473 & 0.3241 \\
    CTGAN & 0.8333 & 0.7990 & 0.3924 & 0.3711 \\
    cWGAN & 0.8443 & 0.7636 & 0.2749 & 0.3325 \\
    CTAB-GAN & 0.8432 & 0.7857 & 0.3519 & 0.3589 \\
    CasTGAN & 0.8557 & 0.8077 & 0.3336 & 0.3866 \\
    \bottomrule
    \end{tabular}
  }
 \end{table}%

\begin{table}[!htb]
  \caption{Cars Dataset Regression}
  \label{tab:app_tstr_cars}
  \centering
  \resizebox{0.70\columnwidth}{!}{
  \begin{tabular}{l|rr}
  \toprule
    {} & {RMSE} & {$R^2$ Score} \\
    \midrule
    Identity & 4330.3428 & 0.7604 \\
    table-GAN & 16737.5294 & -2.0148 \\
    medGAN & 4822.8477 & 0.7491 \\
    CTGAN & 6558.5413 & 0.5327 \\
    cWGAN & 166818.8351 & -319.0963 \\
    CasTGAN & 6396.3703 & 0.5566 \\
    \bottomrule
    \end{tabular}
  }
 \end{table}%

 \begin{table}[!htb]
  \caption{Housing Dataset Regression}
  \label{tab:app_tstr_housing}
  \centering
  \resizebox{0.70\columnwidth}{!}{
  \begin{tabular}{l|rr}
  \toprule
  {} & {RMSE} & {$R^2$ Score} \\
  \midrule
    Identity & 393507.3597 & -0.3696 \\
    table-GAN & 2943381.0066 & -83.1326 \\
    medGAN & 405452.4482 & -0.3845 \\
    CTGAN & 458694.4080 & -0.6533 \\
    cWGAN & 518963.3807 & -1.1054 \\
    CTAB-GAN & 472485.7269 & -0.7284 \\
    CasTGAN & 419848.1947 & -0.4433 \\
    \bottomrule
    \end{tabular}
  }

\end{table}

\break

\subsection{CasTGAN Training Stability}

\begin{figure*}[!htbp]
    \centering
    \begin{subfigure}{0.45\textwidth}
        \includegraphics[width=\textwidth]{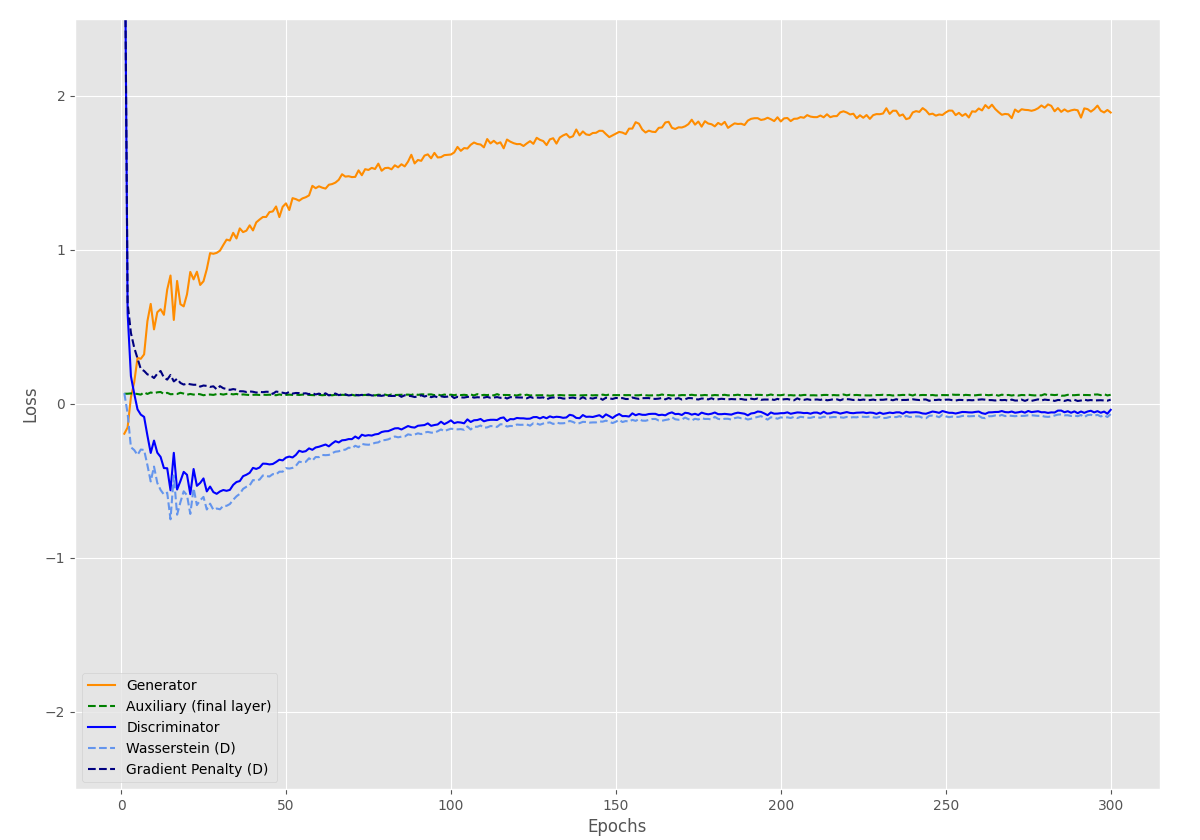}
        \caption{Adult dataset losses}
        \label{fig:loss_plt_adult}
    \end{subfigure}
    \hfill
    \begin{subfigure}{0.45\textwidth}
        \includegraphics[width=\textwidth]{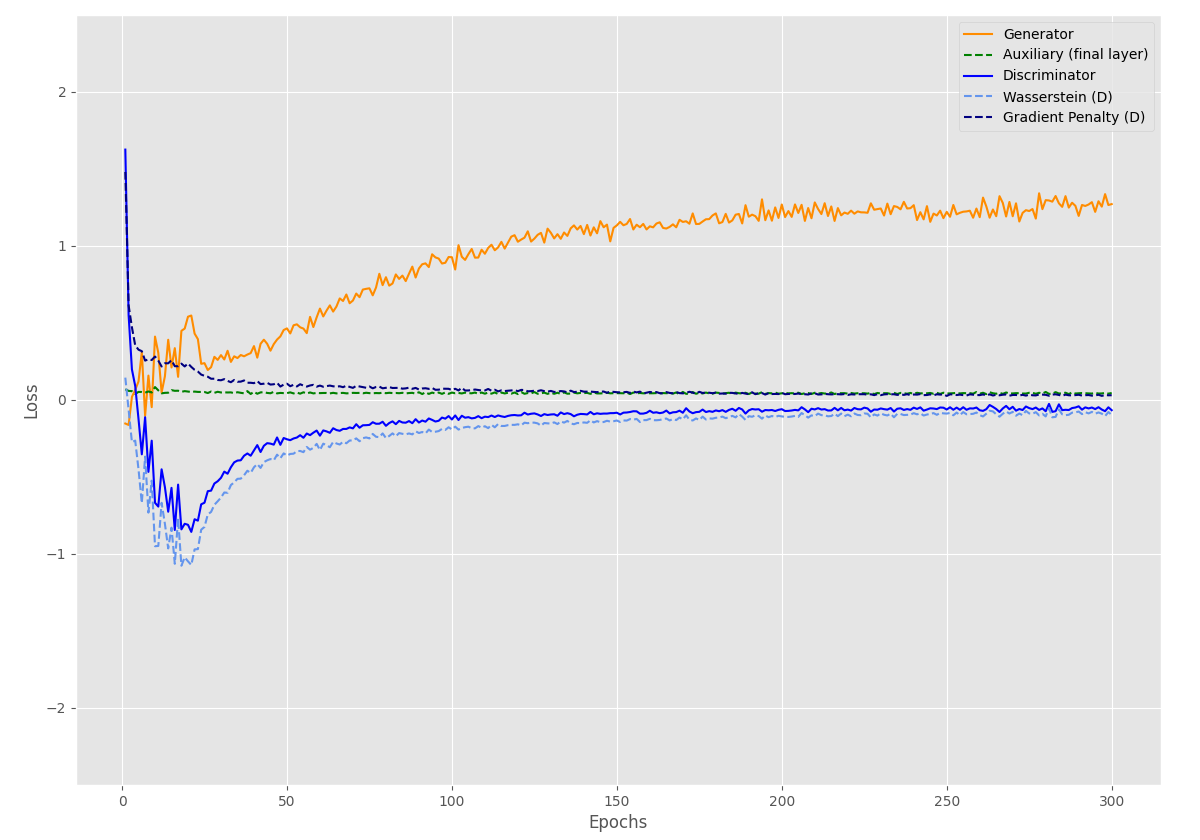}
        \caption{Bank dataset losses}
        \label{fig:loss_plt_bank}
    \end{subfigure}
    \begin{subfigure}{0.45\textwidth}
        \includegraphics[width=\textwidth]{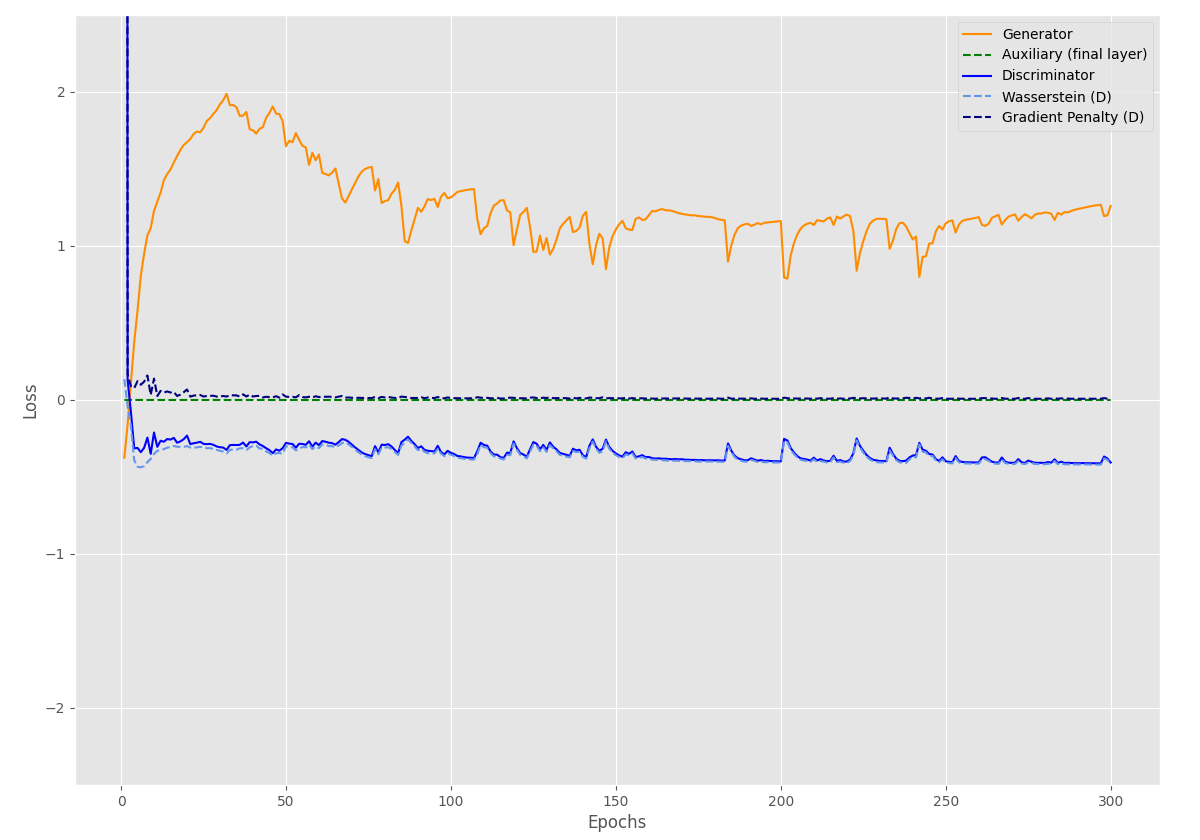}
        \caption{Cars dataset losses}
        \label{fig:loss_plt_cars}
    \end{subfigure}
    \hfill
    \begin{subfigure}{0.45\textwidth}
        \includegraphics[width=\textwidth]{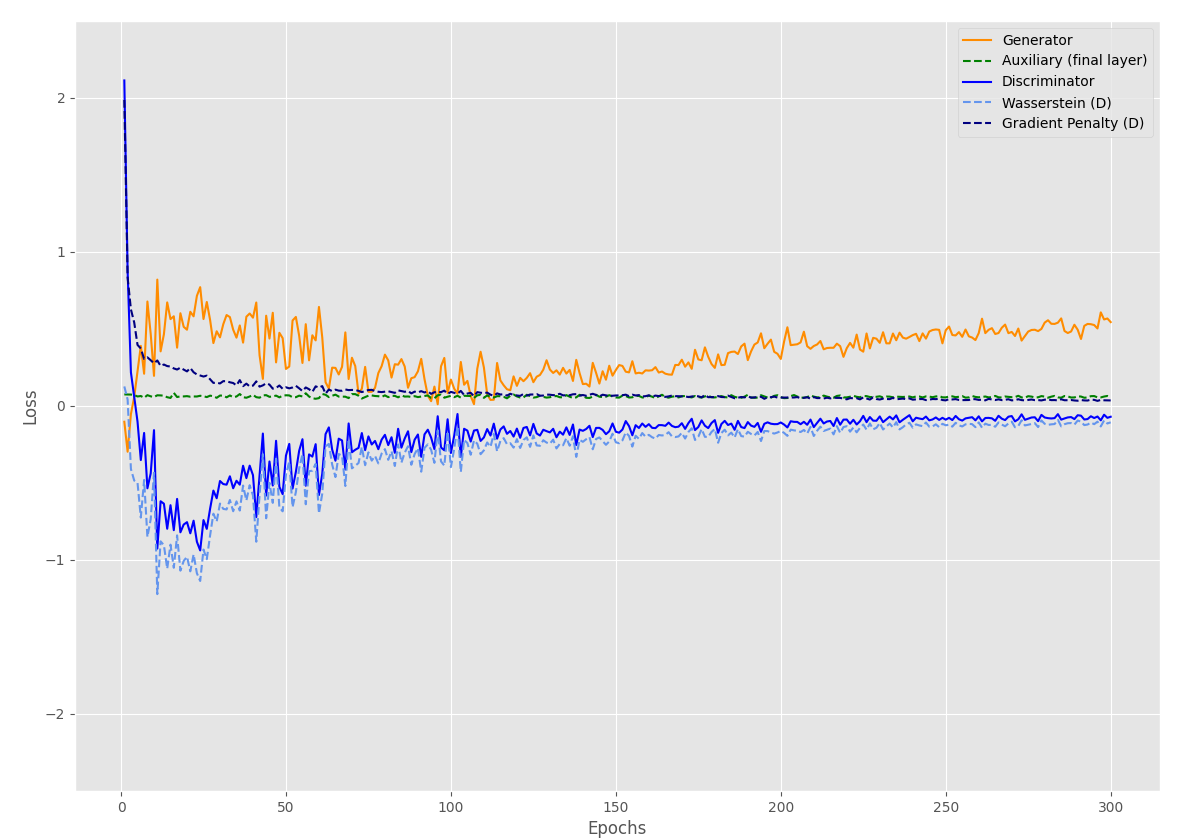}
        \caption{Credit dataset losses}
        \label{fig:loss_plt_credit}
    \end{subfigure}
    \begin{subfigure}{0.45\textwidth}
        \includegraphics[width=\textwidth]{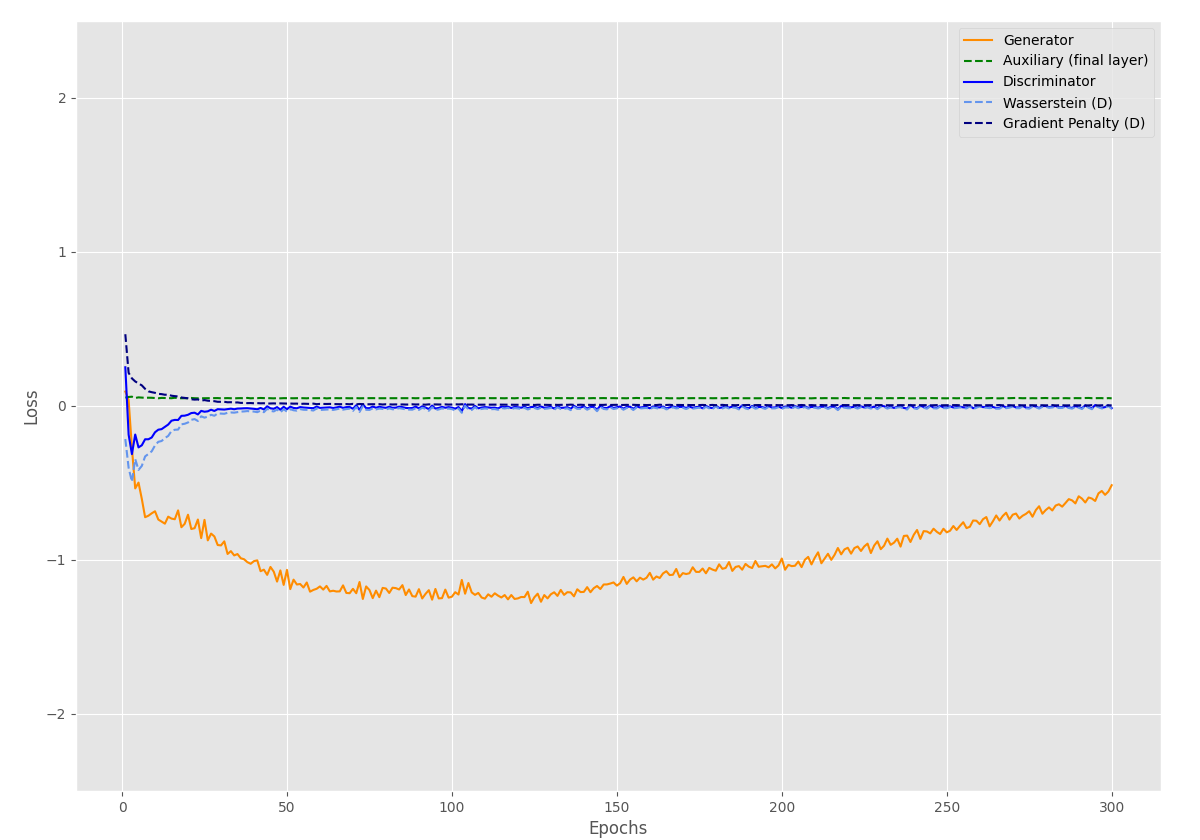}
        \caption{Diabetes dataset losses}
        \label{fig:loss_plt_diabetes}
    \end{subfigure}
    \hfill
    \begin{subfigure}{0.45\textwidth}
        \includegraphics[width=\textwidth]{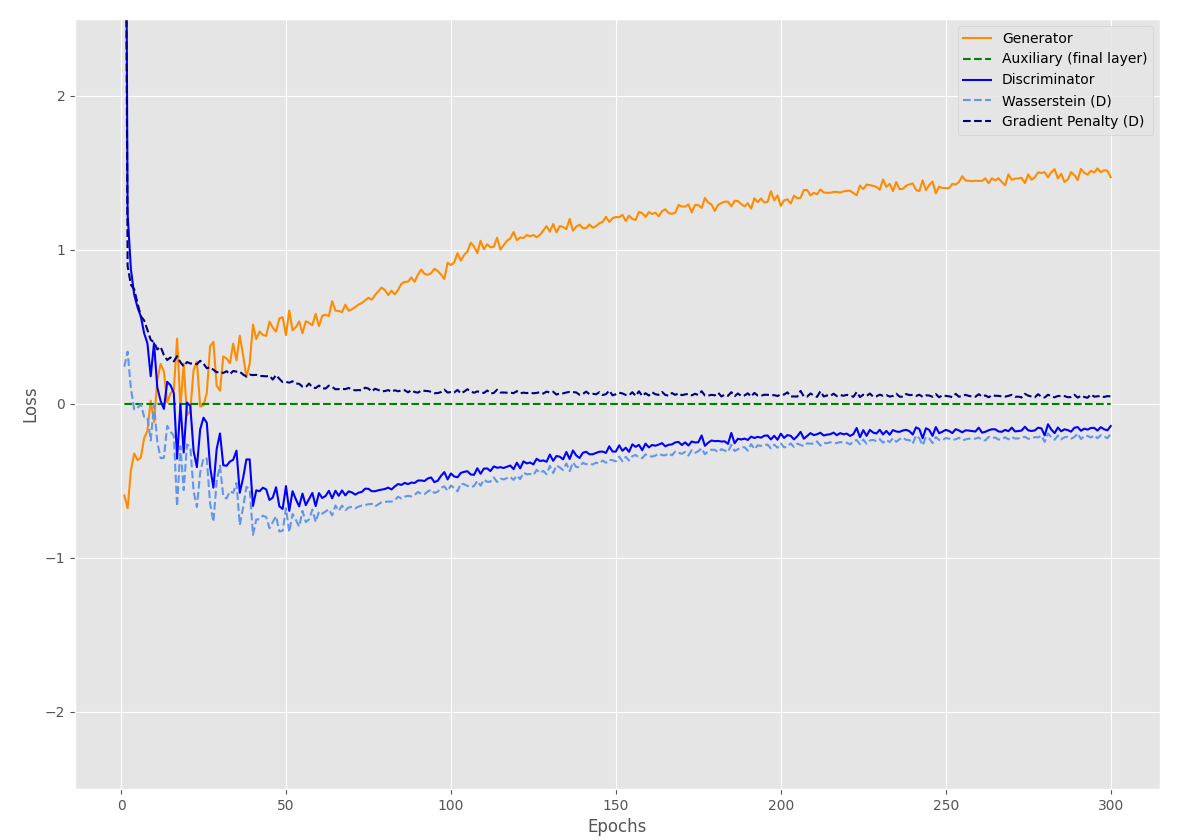}
        \caption{Housing dataset losses}
        \label{fig:loss_plt_housing}
    \end{subfigure}
            
    \caption{CasTGAN training loss plots.}
    \label{fig:loss_plts}
\end{figure*}

\break

\bibliographystyle{IEEEtran}
\bibliography{references.bib}


\begin{IEEEbiography}[{\includegraphics[width=1in,height=1.25in,clip,keepaspectratio]{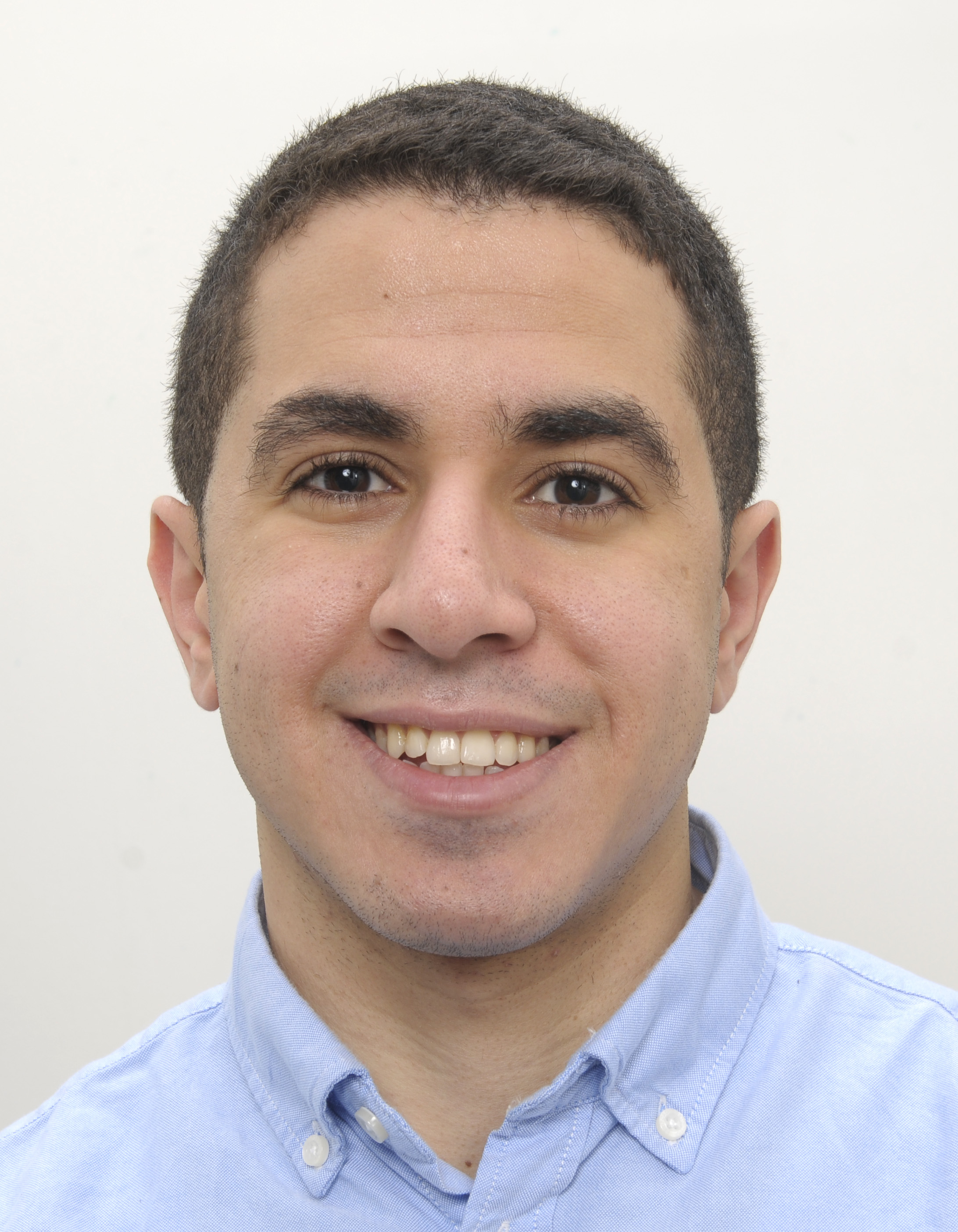}}]{Abdallah Alshantti} received the B.Eng. degree in Mechatronic and Robotic Engineering from the University of Sheffield in 2016, and completed his M.Sc. in Artificial Intelligence at The University of Southampton in 2017. He is currently completing his Ph.D. at the Norwegian University of Science and Technology (NTNU), and has been employed as a Data Scientist at DNB ASA since 2022. His research interests include statistical modelling, machine learning, generative artificial intelligence and anti-money laundering. 
\end{IEEEbiography}

\begin{IEEEbiography}[{\includegraphics[width=1in,height=1.25in,clip,keepaspectratio]{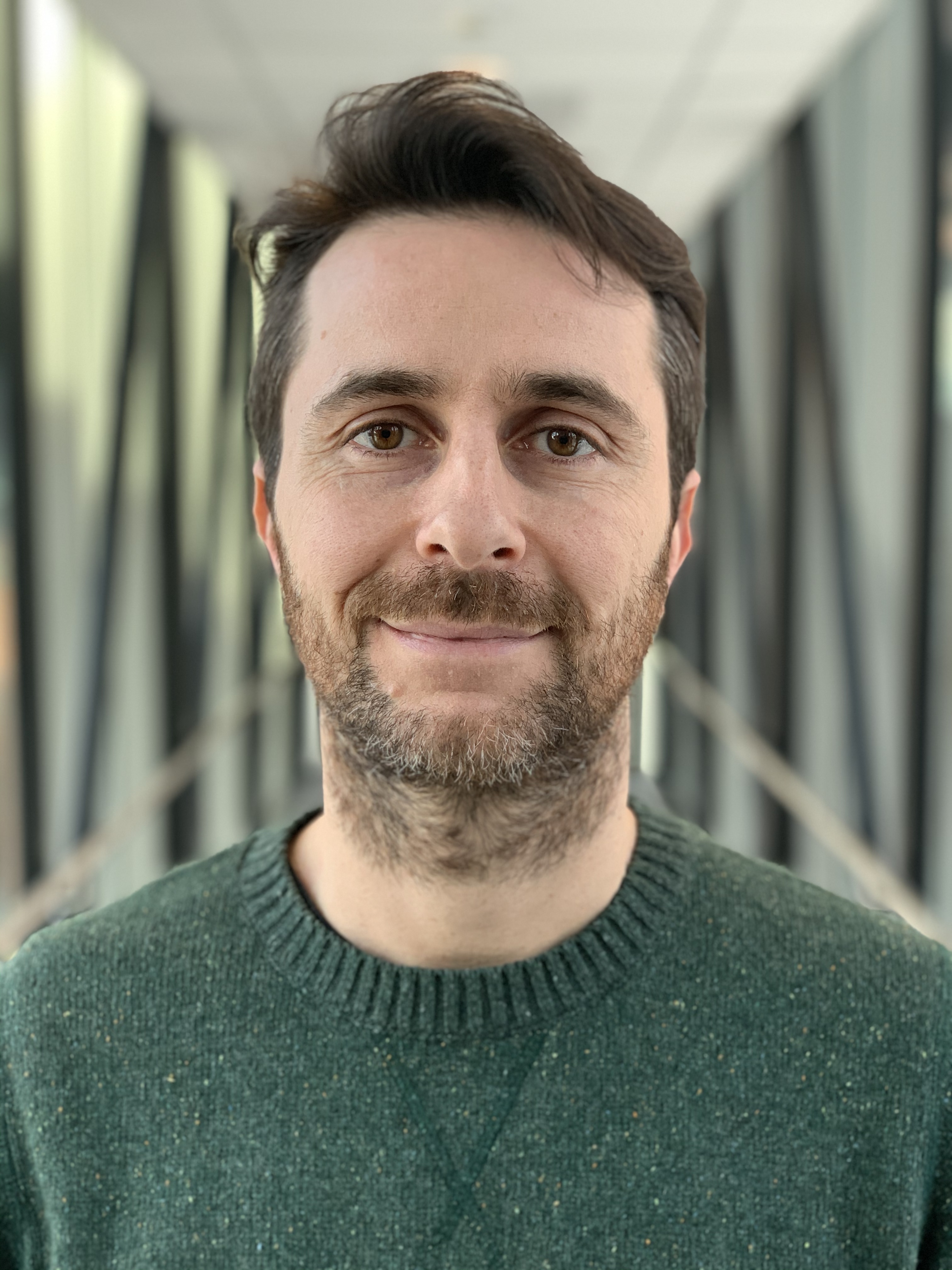}}]{Damiano Varagnolo} received the Dr. Eng. degree in automation engineering and the Ph.D. degree in information engineering from the University of Padova respectively in 2005 and 2011. He worked as a research engineer at Tecnogamma S.p.A., Treviso, Italy during 2006-2007 and visited UC Berkeley as a scholar researcher in 2010. From March 2012 to December 2013 he worked as a post-doctoral scholar at KTH, Royal Institute of Technology, Stockholm. From January 2014 to November 2019 he worked as Lecturer at Luleå University of Technology in Sweden, and then served as Professor at NTNU in Trondheim within the Department of Engineering Cybernetics, with a joint position in University of Padova from 2022. His research interests include statistical learning, distributed optimization, and distributed nonparametric estimation, with a special focus on applications including identification and control for the built environment, learning analytics, and muscular rehabilitation.
\end{IEEEbiography}

\vskip -2\baselineskip plus -1fil
\begin{IEEEbiography}[{\includegraphics[width=0.975in,height=1.3in,clip,keepaspectratio]{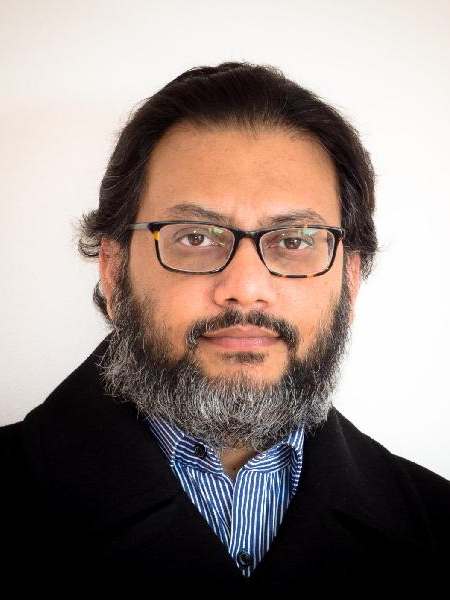}}]{Adil Rasheed} is a Professor in the Department of Engineering Cybernetics at the Norwegian University of Science and Technology. There, he works to advance the development of novel hybrid methods that combine big data, physics-driven modeling, and data-driven modeling in the context of real-time automation and control. In addition, he also holds a part-time Senior Scientist position in the Department of Mathematics and Cybernetics at SINTEF Digital, where he previously served as the leader of the Computational Sciences and Engineering group from 2012 to 2018. His contributions in these roles have been the development and advancement of both the Hybrid Analysis and Modeling and Big Data Cybernetics concepts. Over the course of his career, Rasheed has been the driving force behind numerous projects focused on different aspects of digital twin technology, ranging from autonomous ships to wind energy, aquaculture, drones, business processes, and indoor farming. He is currently leading the Digital Twin and Asset Management related work in the FME Northwind center.
\end{IEEEbiography}

\vskip -2\baselineskip plus -1fil
\begin{IEEEbiography}[{\includegraphics[width=1in,height=1.20in,clip]{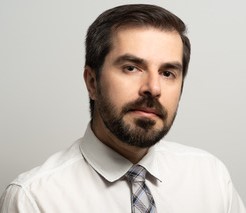}}]{Aria Rahmati} obtained his Ph.D. in 2016 from the University NTNU, specializing in the multifaceted realms of computer vision, machine learning, and medical engineering. Following the successful completion of his doctoral journey,and making a meaningful impact in academia through a series of articles, driven by  passion for practical applications, he transitioned into the industry, where he embarked on the realm of data both as data scientist and and  data engineer. He continues to be a driving force in bridging the gap between research and real-world solutions.
\end{IEEEbiography}

\begin{IEEEbiography}[{\includegraphics[width=1in,height=1.25in,clip,keepaspectratio]{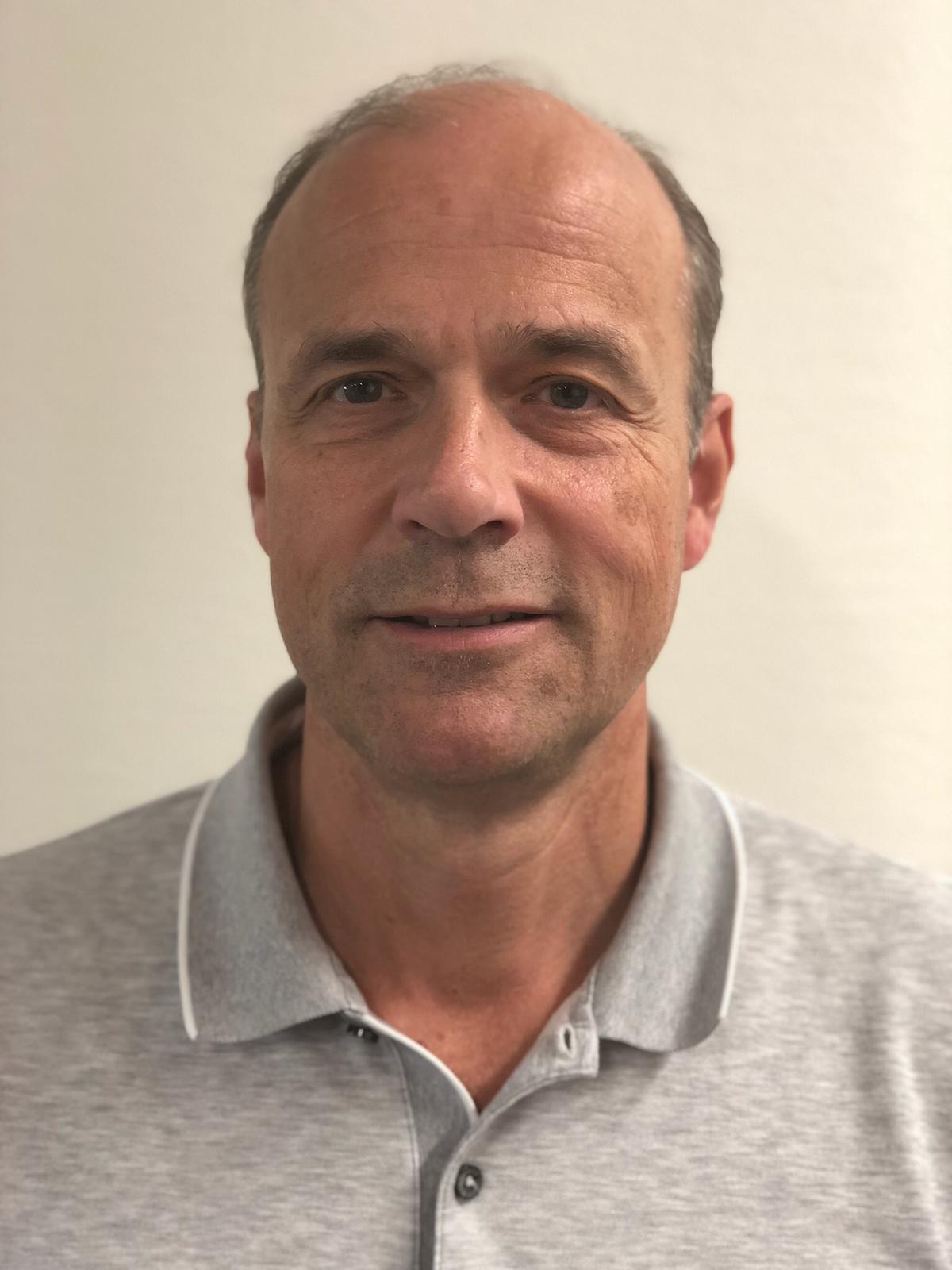}}]{Frank Westad} is an adjunct professor in the Department of Engineering Cybernetics at the Norwegian University of Science and Technology. There, he lectures in multivariate data analysis, machine learning and design of experiments. In addition, he is the lead data scientist at Idletechs AS, a company working with thermal imaging in the metallurgy industry. He has experience from numerous industries such as pharma, food and chemistry, mainly for process modeling with multichannel sensors for real-time process monitoring and prediction.
\end{IEEEbiography}

\EOD

\end{document}